\documentclass{article} % For LaTeX2e
\usepackage{iclr2027_conference,times}

\usepackage{amsmath,amsfonts,bm}

\def\eqref#1{equation~\ref{#1}}
\def\1{\bm{1}}

\DeclareMathAlphabet{\mathsfit}{\encodingdefault}{\sfdefault}{m}{sl}
\SetMathAlphabet{\mathsfit}{bold}{\encodingdefault}{\sfdefault}{bx}{n}

\usepackage{hyperref}
\usepackage{url}

\usepackage{graphicx}
\usepackage{caption}
\usepackage{booktabs}

\title{Diagnosing and Improving Probabilistic \\ Reasoning in Large Language Models}

\author{Huaman Sun, Dingcheng Wang, Jason Hartline \& Jessica Hullman \\
Department of Computer Science\\
Northwestern University\\
Evanston, IL 60201, USA \\
\texttt{\{hmsun,dingchengwang2025\}@u.northwestern.edu \{hartline,jhullman\}@northwestern.edu}}

\usepackage[most]{tcolorbox}
\newtcolorbox{notebox}{
colback=gray!15,
colframe=gray!40,
boxrule=0.5pt,
arc=2pt,
left=6pt,
right=6pt,
top=6pt,
bottom=6pt,
fontupper=\small}

\iclrfinalcopy % Uncomment for camera-ready version, but NOT for submission.
\begin{document}

\maketitle
\lhead{}

\begin{abstract}
Large language models (LLMs) are increasingly proposed as decision assistants who must reason probabilistically from available evidence under explicit decision costs. We propose a decision-theoretic framework that decomposes LLMs' decision loss into two components: forming accurate beliefs from provided evidence and translating those beliefs into actions that optimize a provided utility function. Using a synthetic benchmark with known ground truth, we apply the decomposition to characterize probabilistic reasoning in frontier and open-sourced models. % and identify distinct sources of LLMs' decision loss: errors may arise primarily from inaccurate beliefs, from wrong mapping to actions, or both errors remain substantial. 
%We further investigate whether RL interventions can selectively improve the belief formation and decision-making components of probabilistic reasoning, and how targeting one affects others. %We study four interventions: belief-only training, decision-only training, separate multitask training, and sequential alignment training. 
We further evaluate whether RL interventions targeting beliefs, decisions, or both improve these components across three domains, whether improvements transfer across components and elicitation formats, and whether decision performance can improve without improvement in belief formation. We find that targeting one component of probabilistic reasoning improves the target without necessarily transferring to others, and that jointly targeting belief formation and decision-making improves both but hinges on matched formats between training and evaluation.
%: synthetic data inference, HailFinder weather forecasting, and SimSUM clinical reasoning. We find that single-capacity training redistributes decision loss: it improves its target but does not correspondingly transfer to others. Joint training improves both belief accuracy and action optimality, while its gain largely depends on matched formats between training and evaluation.
\end{abstract}

\section{Introduction}
LLMs are increasingly proposed as decision assistants that can flexibly reason under uncertainty from potentially unstructured evidence and recommend or choose actions. This leads to several fundamental questions about LLMs as probabilistic reasoners: Can they make good inferences and decisions from available evidence? How should such abilities be measured? And, once we can rigorously measure their abilities, can we train models to improve?  
Ideally, domain experts could communicate important context, such as domain-specific preferences and data, and trust that the model's probabilistic reasoning is aligned with domain goals and expertise on new examples. For example, a doctor might use a model to estimate a patient's probability of disease, 
and also want to trust the model to make decisions where the appropriate choice depends on the relative trade-off between false positives and false negatives in the domain.   

Statistical decision theory~\citep{berger1987statistical} provides a natural framework for distinguishing different aspects of probabilistic reasoning under uncertainty. An idealized decision-maker starts with a set of prior beliefs, Bayesian updates those beliefs upon observing new evidence, then chooses the best action under a utility function representing their preferences. In probabilistic reasoning from evidence, beliefs therefore provide an interface between inference and action, where the same beliefs can lead to different choices of action in different decision problems characterized by different utilities. A behavioral decision-maker might experience loss of utility for several reasons relative to this standard: because they arrived at different posterior beliefs than a Bayesian decision-maker would have, or because they failed to optimize their decision under the utility function.

While prior work has explored the calibration of LLMs' token probabilities~\citep{kadavath2022language} and verbalized confidence expressions~\citep{tian2023just,xiong2024can},  understanding probabilistic reasoning in LLMs relative to rational standards is a more nascent aim~\citep{yamin2026agents, yamin2026can, smolin2026beliefs}. Much remains to be understood about LLMs' propensity for two core components of good probabilistic reasoning: the formation of appropriate \textit{probabilistic beliefs} from available evidence and use of those beliefs to make \textit{decisions} under explicitly specified utilities. 

We take inspiration from how fields like cognitive psychology and behavioral economics empirically assess people's probabilistic reasoning ability. 
This controlled approach abstracts away many domain-specific complications, such as ambiguity about the model's preferences or prior, in an attempt to study the underlying belief formation and decision optimization abilities with a minimum of confounding factors.
Here, it is standard to endow beliefs through controlled information structures, such as stated base rates and samples (e.g., \citep{benjamin2019errors,grether1980bayes,holt2009update,kale2020visual}) and utility functions through clearly specified decision scenarios.

We contribute a decision-theoretic framework for assessing core components of LLMs' probabilistic reasoning. We prompt LLMs with binary decision problems for which the Bayesian posterior and optimal action are knowable from the provided information and explicitly provided loss function. This allows us to diagnose departures from rational decision-making by decomposing total decision regret into two sources: belief loss, where the model fails to obtain the Bayesian optimal posterior beliefs, and optimization residual, caused by the model failing to choose the optimal action under the provided utility function.

We first use current models as a testbed for the framework's decomposition, diagnosing belief loss and optimization residuals across five state-of-the-art model families under different inference modes. We find that current LLMs do not share a uniform decomposition signature; rather the distribution of loss across the two sources shifts substantially across task difficulty, reasoning efforts, and model architectures.

We then ask whether the framework can be used as a target for improving LLMs' probabilistic reasoning through post-training. 
We show how our decomposition can be used to derive 
RL-based interventions that target different stages of the belief-to-action pipeline, such as beliefs only, decisions only, or belief-action alignment. %This allows us to assess the extent to which improvements in one aspect of good probabilistic reasoning transfer to other components, and the possibility of training to improve performance on probabilistic reasoning tasks while bypassing the belief formation step entirely.   
We find that belief only training improves its target, but the improvement does not necessarily transfer to better decisions. Decision-only training improves actions without corresponding improvement in beliefs, suggesting that the model may learn a shortcut policy that bypasses the belief formation step. Interventions that jointly target beliefs and actions can improve both components, but their gains depend on the elicitation format. These findings generalize across model families, and transfer to unseen loss functions and decision structures.

\section{Related Work}

Prior work studies how well LLMs can estimate and verbalize confidence in the correctness of their own responses~\citep{kadavath2022language, tian2023just, xiong2024can}. %This work establishes methods for eliciting calibrated uncertainty from black-box language models. However, decision-making often requires a more general probabilistic object: \jessica{not clear why this has to be a different object; that's the question. "However, verbalized probability reports may not relfect the model's latent subjective..."} \huaman{confidence is a small branch within belief -- the belief about the correctness in models' own responses. Decision-making settings, more generally, requires the first-order belief which directly estimate objective posteriors from available evidence. These two concepts are related but not fully interchangeable} \jessica{Again, best to clarify the language; there is nothing that says verbalized probability reports can't reflect underlying beliefs} the model's subjective probability (belief) about an uncertain world state given observed evidence. 
More recent studies examine how LLMs infer such beliefs from available evidence more directly, including whether LLMs can reason about conditional uncertainty from verbalized Bayesian networks~\citep{schrader2024quite}, whether LLMs estimate and update beliefs from in-context evidence in an approximately Bayesian way~\citep{pmlr-v235-falck24a,gupta2025enough}, and how well targeted fine-tuning can improve Bayesian belief updating and transfer performance in user-assistant interactions~\citep{qiu2026bayesian}. We build on work in LLM belief formation, but focus on how these beliefs mediate decisions under specified costs.

A more closely related literature examines misalignment between LLMs' elicited beliefs and actions. \citet{pal2025knowing} %study the relationship between models' static confidence and their behavior in interactive settings. They 
find that well-calibrated uncertainty reports do not necessarily translate into consistent downstream actions. \citet{yamin2026agents} develop decision-theoretic tests for coherence between models' reported beliefs and decisions, assuming  
LLMs possess an internal loss function; % and rationally apply it across decision problems. Their evaluation on clinical diagnosis tasks shows that reported beliefs can deviate from the information revealed by models' decisions. They
\citet{yamin2026can} develop a pipeline for recovering LLMs' internal preferences that best jointly rationalize their elicited beliefs and decisions, finding that LLMs tend to revert to their own preferences rather than faithfully adopting user-specified preferences. \citet{smolin2026beliefs} investigate whether latent belief-like variables can be used to predict models' decisions. Our framework differs by introducing two forms of structure that enable controlled measurements of sources of decision loss. We endow utility functions, % rather than assuming that LLMs act as rational decision-makers who adhere to their internal preferences, 
eliminating ambiguity about what preferences the LLM should act under, and use tasks with known reference posteriors, removing ambiguity about what data-generating model the model should assume. % enables us to decompose total decision loss into errors from belief formation and decision optimization, providing valuable information for understanding and improving LLM decision making under uncertainty.

\section{A Diagnostic Framework for LLM Decision-making}
\label{sec:framework}

\paragraph{Problem setup} 

We introduce a Bayesian decision theoretic framework for assessing LLMs decision-making under uncertainty. Let $x$ denote the observed evidence in the prompt, which may take various forms, from structured tabular data of previous examples to unstructured text such as clinical notes. Let $\mathcal{Y}$ denote the state space, a finite set of possible states of the world. We focus on the binary outcome setting $\mathcal{Y} = \{0, 1\}$, where beliefs can be represented by a scaler probability. A data-generating model defines the true posterior probability 
$p^*(x) =  P(y=1 \mid x)$. An LLM is asked to report a probability $\hat{p}(x)$. Note that we treat $\hat{p}$, the LLM's reported belief which is widely accessible across frontier and open-source models, as an observable intermediary quantity for separating errors related to beliefs from those related to optimization. Our approach does not assume that this report faithfully reflects the LLM's ``true'' belief about the uncertain outcome's distribution. Rather, the goal of our framework is to show how threats to LLMs’ decision-making can be evaluated and used as targets of intervention despite a lack of access to a model’s true beliefs.

Separately, we elicit the LLM's action for an endowed decision problem that specifies a finite action space $\mathcal{A}$ and a loss function $L$. The loss function defines decision quality by assigning a real-valued loss to each combination of action and realized state $\ell:\mathcal{A} \times \mathcal{Y} \rightarrow \mathbb{R}$. In the binary outcome setting, given a probability $p$, the expected loss of action $a$ is 
\[
L(a, p) = \mathbb{E}_{y\sim p}\ell(a, y)=p\ell(a,1)+(1-p)\ell(a,0)
\]
We distinguish three actions: (1) the model's reported action, $\hat{a}$; (2) the optimal action, $a^*=\arg\min_{a \in A} L (a, p^*)$, which minimizes expected loss under the true posterior; and (3) the belief-implied action, $a^{\hat{p}} = \arg\min_{a \in A} L (a, \hat{p})$, which is the action that would be optimal if the model's reported belief were used. Distinguishing these three actions allows us to separate errors in LLMs' probabilistic reasoning from errors in applying a specific loss function.

\paragraph{Regret decomposition} We measure LLMs' decision loss through total regret -- the difference in expected loss under true posterior between the model's reported action and the optimal action:
\[
R_\mathrm{total} = L(\hat{a},p^*) - L(a^*, p^*)
\]
Total regret can be quantitatively decomposed into two parts: belief loss $R_\mathrm{belief}$ and optimization residual $R_\mathrm{opt}$:
\[
R_\mathrm{total} = R_\mathrm{belief} + R_\mathrm{opt}
\]
$R_\mathrm{belief}$ measures the total increase in expected loss from taking the belief-implied action rather than the optimal action:
\[
R_\mathrm{belief} = L (a^{\hat{p}}, p^*) - L (a^*, p^*)
\]
By measuring the decision loss attributable to the reported beliefs differing from the true posterior, $R_\mathrm{belief}$  captures potential sources of loss related to beliefs: an LLM's reported belief may differ from the true posterior because it relies on a different prior, because the LLM failed to extract all decision-relevant information from the evidence, because it updates in a non-Bayesian way, or because it is mis-measured by elicitation.

On the other hand, the optimization residual $R_\mathrm{opt}$ captures the decision consequences of actions being inconsistent with the optimal action implied by reported beliefs:
\[
R_\mathrm{opt} = L (\hat{a},p^*) - L (a^{\hat{p}}, p^*)\]
By measuring the impact of differing belief-implied and reported actions, $R_\mathrm{opt}$ summarizes possible failure of the model to optimize its action to the given utility function, and/or to report its true beliefs. Notably, $R_\mathrm{opt}$ is a signed discrepancy, which can be positive or negative. A negative $R_\mathrm{opt}$ indicates that the LLM's reported action is better than its belief-implied action. This may occur if the model uses a different belief when making a decision from the one it reports in belief elicitation.

\paragraph{Normalization and aggregation} We evaluate LLMs across a set of binary decision problems, where the magnitude of raw regrets depends on the scale of the loss function. Directly averaging across loss functions would overweight those with larger cost scales. Therefore, for a given loss function $L$, we define the maximum possible regret 
\[
M(\ell) = \max_{y \in \mathcal{Y}}\left[\max_{a \in \mathcal{A}}\ell(a,y) - \min_{a \in \mathcal{A}}\ell(a,y)\right]
\]
and normalize the three raw regrets by $M(\ell)$. The normalized total regret and belief loss lies in $[0,1]$, where a larger value indicates greater decision loss. The optimal residual lies in $[-1,1]$, where a negative value indicates the reported action is better than the belief-implied action. 

This normalization removes arbitrary scale variation while preserving the decomposition. However, regret magnitudes remain conditional on the distribution of the true posterior beliefs as well as the loss functions, and should not be read as a reflection of a task's intrinsic difficulty. Numerical comparisons of aggregated regrets should therefore be made within tasks for which the loss functions and the posterior distribution remain fixed.

\section{Diagnosis of state-of-art LLMs}
\label{sec:diagnosis}
Following best practices in studying human probabilistic reasoning and decision-making~\citep{hullman2025underspecified}, we design a controlled synthetic benchmark consisting of decision problems for which the true posterior belief and optimal action can be inferred from the problem description. Compared to giving the model underspecified problems--e.g., without prior labeled evidence or a specified utility function--our approach makes it possible to characterize failure modes without the confound of not knowing how the model is interpreting missing pieces of the problem. We apply the decomposition to a collection of frontier and open-source LLMs across five model families under different reasoning settings. The results can be interpreted as a litmus test of sorts: if a model failures to reason appropriately given well-specified synthetic problems, this warrants caution in expecting it to reason appropriately in more complex, naturalistic settings.

\subsection{Synthetic benchmark construction}

The benchmark consists of controlled decision instances for which the Bayes-optimal posterior belief and action are known by the generating process.  
Each instance presents an LLM with a labeled dataset 
$D_{\mathrm{obs}}=\{(x_i,y_i)\}_{i=1}^n$
and an unlabeled test case $x_{n+1}$, where each $x$ has $k$ binary features and $y\in\{0,1\}$. We vary $k$ in $\{0,1,3,5\}$.
Outcome probability depends only on the number of active features,
$T(x)=\sum_j x_j$, such that 
\[
P(Y=1\mid X=x)=q_{T(x)}.
\]
We construct instances spanning target posteriors 
$p_i^*\in\{1/50,\ldots,49/50\}$
by choosing a monotone sequence
$q_t=\beta_0+\beta_1 t$
satisfying
$q_{T(x_{n+1})}=p_i^*$, where $\beta_1>0$ controls the strength of the association between active features and outcome, and $\beta_0$ gives the baseline probability. For each instance, parameters are chosen so that $q_{T(x_{n+1})}= p^*_{i}$.

We construct $D_{\mathrm{obs}}$ so that its empirical conditional frequencies equal the specified $q_t$.
For each value of $T(x)$, we include 50 observations with the corresponding proportion of positive outcomes, with feature vectors are sampled uniformly conditional on $T(x)$. For each $D_{\mathrm{obs}}$, we construct five instances where we random the orderings of observations to reduce sensitivity to presentation order.

For each instance, we prompt the model separately for its belief under the quadratic scoring rule, and its decisions across nine threshold losses. We provide prompt templates in Appendix~\ref{app:synthetic_evaluation}.

\subsection{Results} 

\begin{figure}[htb]
    \centering
    \includegraphics[scale=0.85]
    {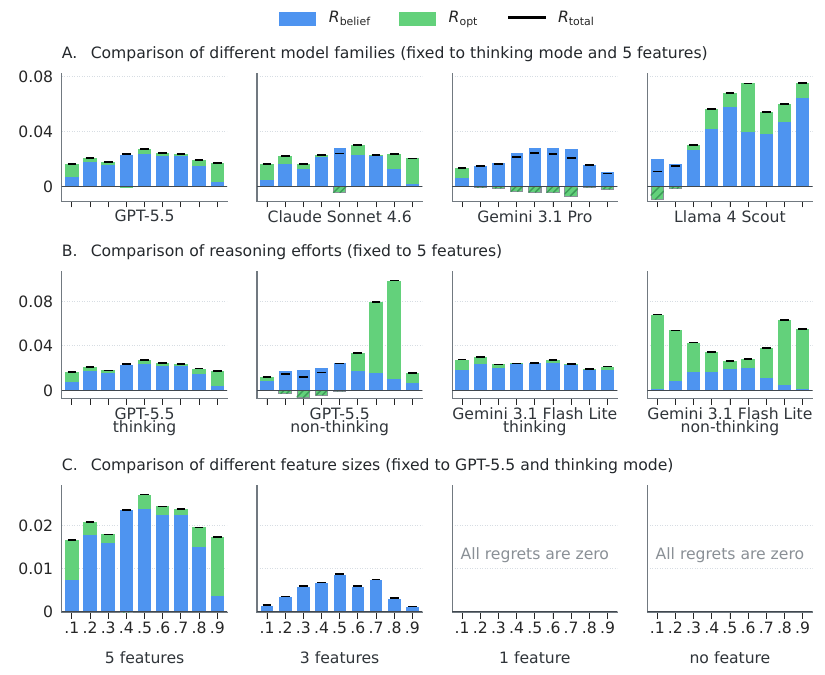}
    \caption{Normalized regret decomposition on the synthetic benchmark for representative models.
Rows compare model families (A), reasoning efforts (B), and feature sizes (C).
Each bar decomposes total regret into belief loss (blue) and optimization residual (green); black marks indicate the total regret.
Note the different y-axis scale in Row C.}
    \label{fig:main-decomposition}
\end{figure}

\paragraph{The magnitude and relative contributions of the decision loss components vary substantially depending on models, reasoning efforts, and task difficulty.} Figure~\ref{fig:main-decomposition} summarizes results with featured models; we provide full results across ten model-reasoning settings in Appendix~\ref{app:synthetic_results}. %Our results suggest that belief accuracy and decision optimality should not be treated interchangeably. The decomposition shifts with the model, feature size, reasoning mode, and loss function. 

For simple statistical reasoning tasks, such as learning from observations with zero or one feature, we find that frontier models with thinking mode on exhibit perfect performance in both reporting beliefs and making decisions. GPT-5.5 with medium reasoning has zero regret at every threshold, as do Gemini 3 Flash, Gemini 3.1 Pro, and Claude Sonnet 4.6 at high reasoning effort. 

For models that are optimal in the simple settings, as inference becomes more difficult, the primary source of regret that emerges is belief loss. In Row~C, regret first appears at $d=3$ and increases at $d=5$, primarily through belief loss. At $d=5$, Row~A shows a similar decomposition dominated by belief loss for Claude Sonnet 4.6 and Gemini 3.1 Pro under high reasoning effort. 

In contrast, for models that fail to optimize given utility functions under reported beliefs in fewer feature settings, the optimization residual dominates total regret. GPT-5.5 with no reasoning already has large residuals at $d=0$ and $d=1$. At $d=5$ (Row~B), its $R_\mathrm{opt}$ at $\tau = 0.7$ and $0.8$ are substantially larger without reasoning than under medium reasoning. Gemini 3.1 Flash Lite shows a similar shift in $R_\mathrm{opt}$ when its reasoning changes from high to minimal.

For observations where the model's decisions implied a single switch point, we exploit an equivalence between 
proper scoring rules for beliefs on binary states and mixtures of elementary threshold losses~\citep{gneiting2007strictly} to compare LLMs' reported beliefs to their decision-revealed beliefs, which for a rational decision-maker will be equivalent. %This tests for equivalence between belief representations obtained through prompting with a proper scoring rule versus inferred from a model's repeated decisions. 
We find that models can generally estimate beliefs with reasonable accuracy, but may fail to map the belief to making cost-sensitive decisions. We provide details for this supplementary analysis in Appendix~\ref{app:revealed_beliefs}.

\paragraph{Comparisons to further disambiguate contributors to belief loss or optimization residuals.} Our framework captures a broad distinction between error sources arising from belief related and optimization related failures. Additional experiments can be used to further disambiguate sources of loss. We demonstrate with two supplementary experiments: 1) evaluating wrong inferences about the data-generating model as a source of belief loss; and 2) distinguishing genuine optimization errors from belief-action misalignment.

Our synthetic benchmark problems do not directly describe the form of the data-generating model in the prompt. Thus, the model must infer a form by observing the evidence. Belief loss may therefore reflect the model making different assumptions about the relationship between features and outcomes than hold in the true data-generating model. In Appendix~\ref{app:posterior_empirical_frequency}, we report how the diagnosis results on the synthetic benchmark differ when true posteriors is defined by an alternative rule: conditional on specific feature identities in the same $D_{\mathrm{obs}}$. We find that belief loss, along with total regret, reduces across tested scenarios, suggesting that the model may have assumed a different form of the data-generating model as a contributor to decision loss.

To answer whether measured optimization residuals mainly reflect genuine optimization errors or belief-action misalignment, we evaluate the decision consequences when a model is directly provided with the beliefs, and must only choose the best action. We test GPT-5.5 in both thinking and non-thinking modes. We find that with medium thinking effort, the model reliably chooses belief-implied optimal actions when probabilities are explicitly available, which suggest that optimization residuals in independent evaluation may mainly reflect belief-action misalignment. On the other hand, we find that without thinking, belief–action agreement remains between 87.0\% and 89.9\% across the tested conditions, which may reflect genuine difficulty optimizing the given utility functions. We provide details and full results in Appendix~\ref{app:distinguish_R_opt}.

\section{Improving Probabilistic Reasoning with Targeted RL Post-training}

A useful diagnostic framework can also be used to improve models' reasoning capacities. %Our decomposition distinguishes distinct sources of LLMs' decision regret. %We evaluate how targeting each through RL-based post-training impacts  leads to independent improvements in that capability versus changes to other capabilities or overall belief-decision alignment. 
We evaluate the extent to which targeting RL reward interventions to particular capacities of probabilistic reasoning can selectively improve belief formation and decision optimization versus have spillover effects, and whether targeting good decisions can improve performance without improving belief formation. We study four interventions: belief-only training, decision-only training, separate multitask training, and sequential alignment training.

\paragraph{Belief-only training (\textit{B})}rewards the model for reporting accurate beliefs from observed evidence, with no supervision on decisions. Given a reported belief $\hat{p}$ and true posterior $p^*$, we define the belief reward as 
\[
r_\mathrm{B}(\hat{p}, p^*) = 1 - (\hat{p} - p^*)^2
\]
The belief reward lies in $[0,1]$, and is maximized when the reported belief equals the true posterior.
We use belief-only training to test the extent to which improving belief formation alone can lead to better decisions. It also allows us to evaluate whether gains from improving beliefs can be canceled by weaker belief-action alignment, reflected in increased optimization residuals after post-training.

\paragraph{Decision-only training (\textit{D})} directly rewards the model for choosing the optimal action under the true posterior and an explicit loss function, with no feedback on belief accuracy. For model's reported action $\hat{a}$, true posterior $p^*$, and loss function$\ell$, we define the decision reward as
\[
r_\mathrm{D}(\hat{a}, p^*, \ell) = 1 - R_\mathrm{total}^\mathrm{norm}(\hat{a}, p^*, \ell)
\]
The decision reward lies in $[0,1]$, and is maximized when the reported action equals the optimal action.
Targeting the final action during post-training may result in the model inferring more accurate beliefs, better mapping between beliefs, or simply learning an action-specific policy without forming an accurate, reusable belief representation that can be verbally elicited. Comparing the effects of this intervention on total regret and belief loss examines whether improved actions compensate for larger belief errors. 

\paragraph{Separate multitask training (\textit{B/D})} combines belief- and decision-only training by randomly allocating half of the training examples to each component. Belief examples receive $r_\mathrm{B}$, while decision examples receive $r_\mathrm{D}$. Decisions are elicited independently rather than conditioned on reported beliefs. This intervention tests whether separately rewarding belief formation and decision optimization is sufficient to improve both without creating competition between them. It also provides a baseline to distinguish whether improvement in sequential training (described below) comes from receiving both forms of supervision, or from rewarding alignment of downstream actions with reported beliefs. 

\paragraph{Sequential alignment training (\textit{B+A})} trains both accurate belief reporting and alignment between selected decisions and reported beliefs  within the same prompting session. The model is first asked to report its belief and receives $r_\mathrm{B}$. In a second turn, it observes the loss function, selects an action with access to its previous reported belief, and receives an alignment reward. We follow the general implementation of turn-level credit assignment in multi-turn RL~\citep{zeng2025reinforcing}. We assign the target-specific reward to each turn's response, rather than providing an aggregated conversation-level signal.% that obscures whether belief reporting or action selection earns the reward. 
Given a belief-implied action $a^{\hat{p}}$ and the reported belief $\hat{p}$, we define the alignment reward as
\[
r_\mathrm{A}(\hat{a}, \hat{p}, \ell) 
= 1 - \frac{L(\hat{a}, \hat{p}) 
- L(a^{\hat{p}},\hat{p})}{\mathrm{M}(\ell)}
\]
It lies in $[0,1]$, and is optimized when the reported action is optimal under the reported belief.
This tests whether explicitly encouraging actions aligned with reported beliefs improves decisions and whether it transfers robustly across formats. 

%Each of the four interventions targets a different capacity implicated in Bayesian decision-making. We compare the effects on the decomposition to examine both improvements in the targeted source of loss and whether these improvements transfer to other capacities in the belief-to-action pipeline, or redistribute decision loss elsewhere.

\subsection{Experiments}
\label{sec:experiment}

We evaluate the four RL interventions across three domains across a spectrum from highly controlled to more naturalistic: synthetic data inference, weather forecasting, and clinical reasoning. The decision environments share the same nine endowed binary loss functions but differ in the structure of the evidence, requisite prior knowledge, and the structure of the data-generating model. 

\paragraph{Tasks} The synthetic task provides a controlled environment where the true posterior can be inferred exactly from provided tabular observations. Each instance contains 300 previous observations with five binary features and a binary outcome. Given a test case, the model must estimate its outcome probability and decide whether to assign a positive label under a specified loss function. %Because both the true posterior belief and evidence-generating process are known, the synthetic task endows controlled beliefs to LLMs without prior knowledge contamination, providing a clean setting for evaluating the effects of post-training interventions.

The weather forecasting task requires decisions from structured text input, where the ground truth is given by HailFinder~\citep{abramson1996hailfinder}, an expert-designed Bayesian network with 56 variables for severe-weather forecasting. We generate instances by sampling the seven upstream atmospheric variables and marginalizing over all others. We render the selected variables as short natural-language descriptions. The model must estimate the probability of significant or severe hail, and decide whether to issue a warning under a provided loss function.

The clinical task further tests probabilistic reasoning with unstructured evidence that resembles medical diagnosis scenarios, but with the unique property that true posteriors are also available. SimSUM~\citep{rabaey2025simsum} provides structured symptom records generated from an expert-defined casual model, as well as the corresponding unstructured AI-generated free-text clinical notes with expert verification. %Given the structured symptoms, we derive posteriors from the known causal models. %This makes SimSUM a good task to evaluate LLM decision-making with unstructured evidence, where true posteriors conditional on evidence itself are rarely accessible. 
We use the clinical notes as evidence and ask the model to estimate the patient's probability of infectious respiratory conditions. The model also decides whether to escalate the patient for additional evaluation given costs. 

Full data construction, split sizes, and example inputs are provided in Appendix~\ref{app:synthetic},~\ref{app:hail}, and~\ref{app:simsum}.

\paragraph{Decision problems} Across all three tasks, we use the same nine binary loss functions for training and evaluation. Correct actions receive zero cost. False-positive and false-negative actions receive costs of $c_\mathrm{FP}$ and $c_\mathrm{FN}$, respectively. We vary the relative costs while fixing $c_\mathrm{FP} + c_\mathrm{FN} = 10$, resulting in uniformly distributed optimal action thresholds $\tau = \frac{c_\mathrm{FP}}{c_\mathrm{FP} + c_\mathrm{FN}} \in [0.1,0.2,\dots,0.9]$.

\paragraph{Model and training control} We use Qwen3-8B as the base model for all post-training experiments. For each strategy and task, we train a separate LoRA adapter from the same base checkpoint, and optimize using GRPO with eight sampled completions per prompt. Within each task, all four interventions use the same training split, base-model initialization, optimizer configuration, and fixed training schedule. We train five epochs on the synthetic task and two epochs on the weather forecasting and clinical task. We monitor training progress on a held-out validation split and confirm that each intervention stabilized by the final checkpoint. 

\paragraph{Evaluation} We evaluate each post-training intervention using normalized total regret, belief loss and optimization residual as defined in Section~\ref{sec:framework}. We also evaluate two additional metrics--MSE calculated between the reported belief and true posterior; and agreement between the belief-implied and reported action--to corroborate our claims about the effectiveness of post-training. We report the averaged differences in metrics relative to the base model, with bootstrap 95\% confidence intervals.
\[\mathrm{Belief\_MSE} = (\hat{p} - p^*)^2\]
\[
\mathrm{Agreement}(\hat{p}, \hat{a})
=
\mathbf{1}\!\left\{
\hat{a} \in \arg\min_{a \in \mathcal{A}} L(a, \hat{p})
\right\}
\]
We evaluate results using two elicitation formats. In the \textit{independent format}, we elicit beliefs and actions in separate conversations. This enables us to assess whether separate belief and action readouts appear to share the same belief representation. In the \textit{two-turn format}, the model chooses actions under specified costs within the same conversation, such that the reported belief remains in context. This enables us to assess how the access to reported beliefs impacts belief-action alignment.

\subsection{Results}

\begin{figure}[htb]
    \centering
    \includegraphics[width=0.95\textwidth]{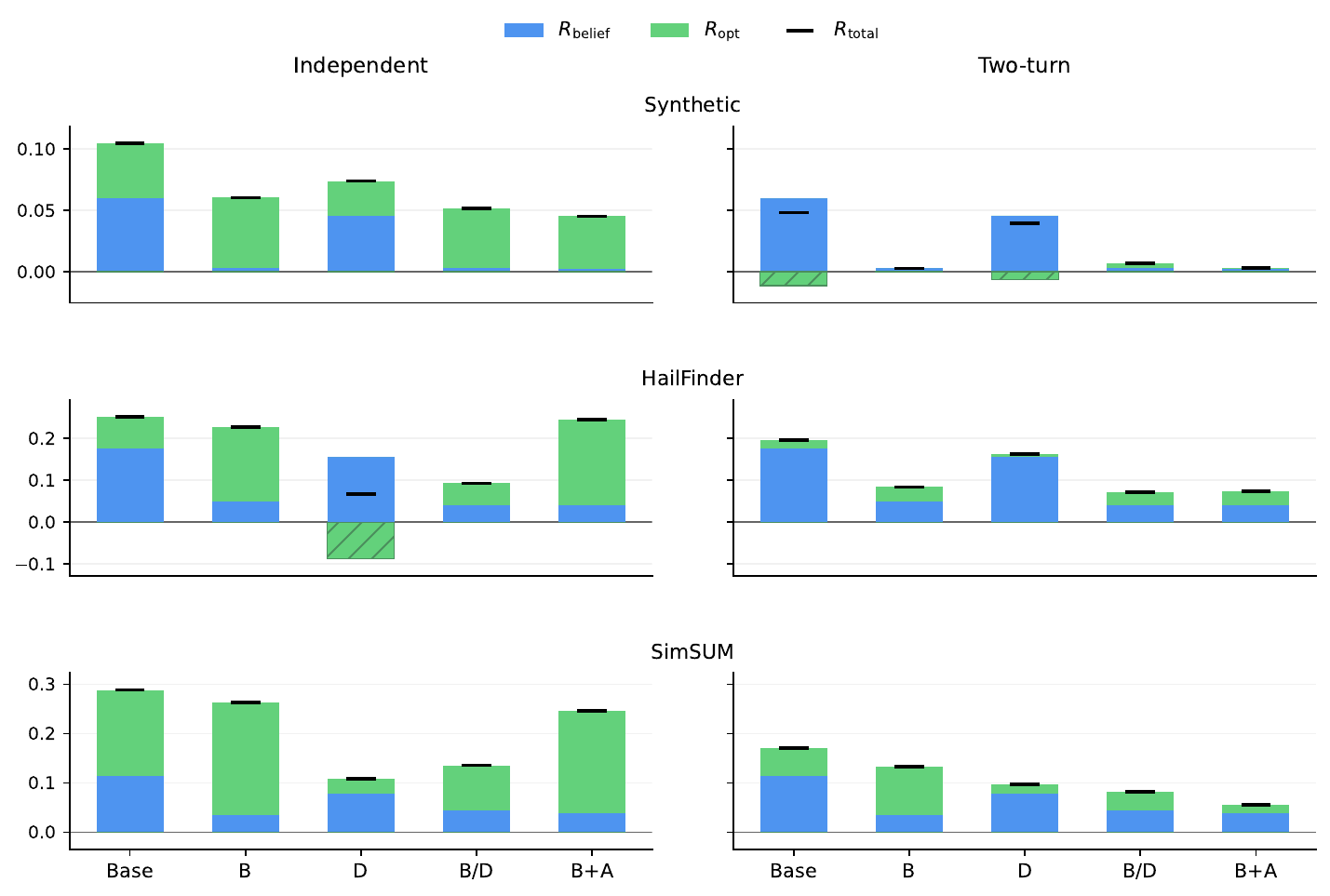}
    \caption{Aggregated regret decomposition of the base model (Qwen3-8B) and after the four interventions. Rows show synthetic, weather and clinical tasks; and columns show independent and two-turn evaluation.}
    \label{fig:result_abs}
\end{figure}

\begin{figure}[htb]
    \centering
    \includegraphics[width=\textwidth]{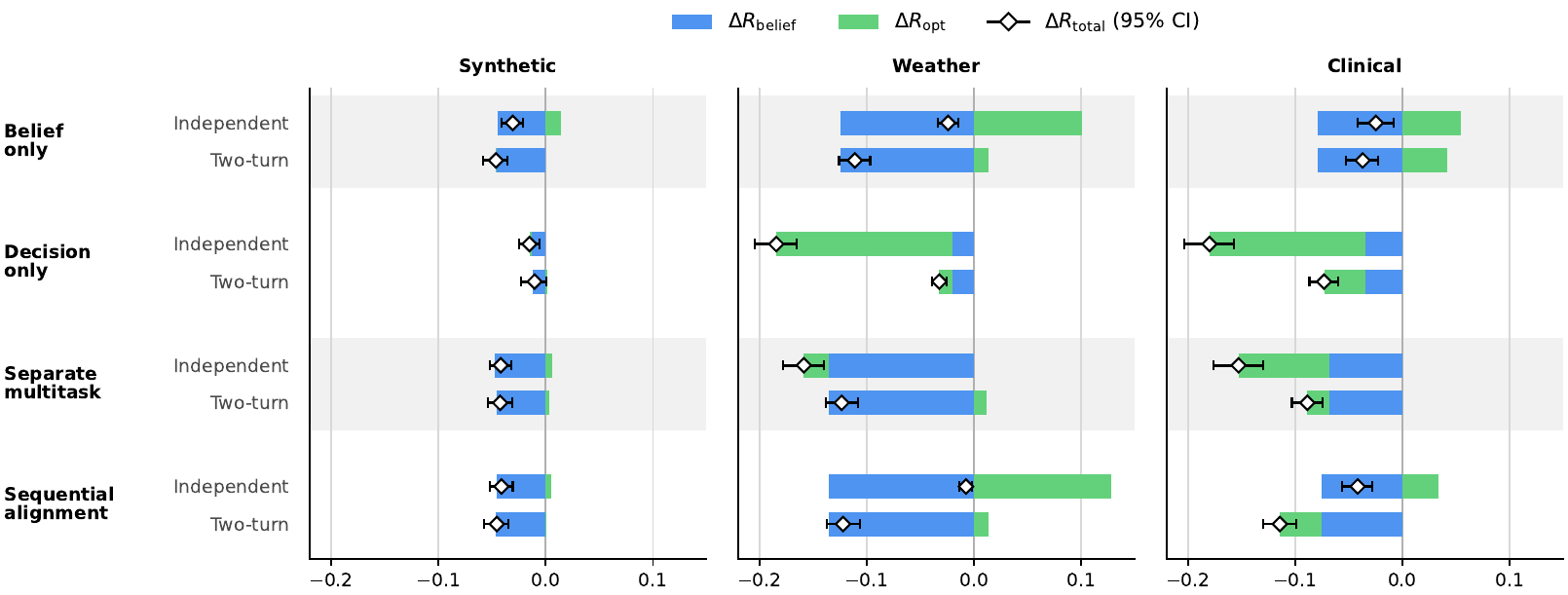}
    \caption{Averaged differences from the base model (Qwen3-8B) in normalized total regret, belief loss, and optimization residual. Rows show the four interventions (belief-only, decision-only, separate multitask, and sequential alignment training); and subrows distinguish between the independent and two-turn evaluation formats. Values below zero indicate a reduction (improvement) in the corresponding term relative to the base model.}
    \label{fig:result_diff}
\end{figure}

Figure~\ref{fig:result_abs} shows the aggregated performance of the base model and after each of the four interventions. Figure~\ref{fig:result_diff} further illustrates the effects of each intervention on the regret decomposition. We provide numerical results with 95\% bootstrap confidence intervals in Appendix~\ref{app:results}.

\paragraph{Loss magnitudes vary with task and evaluation format} Across all three tasks, the base model exhibits nontrivial belief loss and optimization residual, but the breakdown varies by task and format. In the independent format, the synthetic task shows a nearly balanced decomposition, with substantial loss from both sources. In contrast, belief loss is the major source of decision loss in the weather forecasting task, whereas optimization residual accounts for a larger proportion of total regret in the clinical task. This indicates that even for the same model, there does not exist a task-independent pattern of decision loss.

The two-turn evaluation on the base model reveals that keeping reported beliefs in context reduces total regret while leaving belief loss unchanged. Compared to independent action elicitation, this suggests that explicit belief reporting can improve LLMs' decision-making but can not correct losses from inaccurate probability estimates. Because the belief prompt is identical across evaluation formats, the reduction is attributable to changes in the belief-to-action mapping. In the synthetic task, the aggregated optimization residual becomes negative, suggesting that reported actions sometimes compensate for belief errors. In the weather and clinical task, the optimization residual remains positive but is substantially smaller.

\paragraph{Belief only training improves its target, while the improvement does not necessarily transfer to better decisions.} Belief-only training consistently reduces belief loss across controlled, structured and unstructured evidence. However, improvements from targeting belief accuracy do not necessarily transfer to comparable reductions in decision loss. Under independent evaluation, reductions in total regret are substantially smaller that improvements in belief loss. The two-turn format, which provides reported beliefs in context, can encourage belief-aligned actions and make improved beliefs more useful for final decisions. In all three tasks, belief-only training reduces total regret under two-turn evaluation.

\paragraph{Decision-only training improves actions while bypassing the belief formation step.} Across tasks, decision only training consistently reduce total regrets, particularly in the weather forecasting and clinical tasks. However, belief loss decreases only slightly from the base model, suggesting that directly minimizing total regret does not improve beliefs to the same extent as actions. Instead, the model may learn a shortcut policy that bypasses the explicit belief formation step. The two-turn evaluation provides further evidence: when the model first reports its beliefs and then selects actions, the benefits of decision only training are much smaller across the three tasks.

\paragraph{Separate multitask and sequential alignment training improve both components, but in different forms.}

Both joint interventions use the belief supervision and consequently reduce belief loss similarly to belief-only training, even though separate multitask training allocates only half of its training examples to beliefs. However, their effects on actions differ across evaluation formats. 

Separate multitask training consistently reduces total regret, with gains balanced across both sources of loss. Its gains are also robust to evaluation format. This suggest that the improved capabilities remain accessible whether beliefs and actions are elicited independently or in a two-turn conversation.

Sequential alignment training is most effective when evaluation matches the trained belief-to-action pipeline. It is among the strongest interventions under two-turn evaluation, but its gains become smaller under independent evaluation because the model cannot access its reported beliefs when selecting actions.

\paragraph{Interventions effects generalize to different model families.} We replicate post-training experiments on Llama3.1-8B-Instruct. Overall, the intervention effects are consistent with those observed from Qwen3-8B. Belief-only and decision-only training improve their respective targets, without corresponding improvements in the other components. Separate multitask and sequential alignment training improve both components, but the effects depend on evaluation formats. One model-specific difference is the effectiveness of belief supervision in sequential alignment training. For Llama3.1-8B-Instruct, adding belief–action alignment supervision yields smaller improvements in belief accuracy than belief-only training. For Qwen3-8B, sequential alignment training maintains a similar level of belief improvement to belief-only training. We provide detailed results in Appendix~\ref{app:replication}.

\paragraph{Improvements transfer to untrained loss functions and decision structures.} We conduct supplementary experiments to test whether intervention outcomes transfer to untrained decision problems. We first test whether training on a subset of the thresholds $\tau \in \{0.2, 0.4, 0.6, 0.8\}$ generalize to unseen thresholds $\tau \in \{0.1, 0.3, 0.5, 0.7, 0.9\}$. We separately train each intervention with the subset of thresholds for the synthetic task on Qwen3-8B. We find that the interventions generalize well to the held-out thresholds, and the main findings continue to hold.

We further test whether training on binary threshold decisions transfers to more complex decision problems without additional training. We conduct experiments on the clinical task across two base models. We introduce a defer option for intermediate probabilities, which results in two thresholds for selecting the optimal actions. We evaluate six pairs of lower and upper thresholds which cover small to large defer regions. We find that the improvements from decision-only and separate multitask training transfer to three-action settings. The effects of sequential alignment training also transfer but remain dependent on the evaluation format. We provide full results in Appendix~\ref{app:utility_generalization}.

\section{Discussion}

In this work, we develop a decision-theoretic framework that decomposes LLMs' decision loss into two components: errors in forming beliefs from available evidence and errors in translating those beliefs into optimal actions under specified utilities. We apply the decomposition on a controlled synthetic task to characterize probabilistic reasoning in state-of-art frontier and open-sourced models. These results can provide valuable information about what is reasonable to expect in more complex scenarios: if a model cannot optimize its behavior when provided with a well-defined decision problem, it is unreasonable to expect that it will form accurate beliefs and make utility-optimal decisions in more complex and uncontrolled settings. We further investigate whether RL interventions can target different components on the belief-to-action pipeline, and whether improvement transfer across components and elicitation formats. We find that targeting one component of probabilistic reasoning improves the target without necessarily transferring to others, and that jointly targeting belief formation and decision-making improves both but hinges on matched formats between
training and evaluation.

How well LLMs can reason probabilistically from evidence impacts their trustworthiness across a number of domains where they are currently turned to as decision assistants. Our work proposes a foundational framework for diagnosing and improving distinct sources of loss in LLM probabilistic reasoning. 
In doing so, we address several challenges in rigorously diagnosing reasoning failures, including the potential for confounds due to the model not having access to all information used in defining true posteriors and optimal actions or acting under a different set of preferences than intended. 
By endowing information structures and explicit decision problems, our work provides tools for overcoming some of these challenges. Although we choose to focus on controlled tasks to avoid introducing additional ambiguity in estimating ground truth, our framework and intervention methods can be applied to more realistic scenarios where known ground truth can still be established.
%However, as we show in Appendix~\ref{app:posterior_empirical_frequency}, our procedure cannot fully remove ambiguity about the structure of the data-generating model. Studying inductive biases in model class inference is a fruitful area for future work. 

Other opportunities to extend our results consider different decision strategies, elicitation techniques, and model families. We evaluate decision loss against optimal Bayesian decision-making, however, alternative standards one might be interested in evaluating and training against include forms of robustly optimal decisions.  
Next, our decomposition relies on reported beliefs, which may not faithfully reflect internal representations due to elicitation distortions. Future work could incorporating probing methods to separate out elicitation loss from belief formation loss.
In addition, our experiments focus on binary states and decision problems, although the framework is not limited to binary scenarios. The decision-theoretic decomposition applies to any finite state space and action space. Developing controlled benchmarks with multiclass outcomes and larger action spaces would broaden its application to understanding LLMs probabilistic reasoning.
%Finally, our RL experiments are conducted on a single base model. Examining them across model families and scales would help distinguish generalizable effects from model-specific ones. 

Future work could also systemically test whether intervention gains transfer beyond the training settings, helping distinguish generalizable improvements from learning task-specific mappings during training. For example, belief interventions targeting a specific inference rule could be evaluated on datasets with similar statistical structures but in different domains. Our generalization tests on unseen thresholds and three-action space provides examples of successful transfer across decision structures. But systematically training and evaluating generalizable probabilistic reasoning may require more sophisticated reward design and more comprehensive data construction across a variety of decision problems, which is a fruitful area for future work.

\subsubsection*{Acknowledgments}
This work used GPU computing resources at DeltaAI from the \href{https://access-ci.org/}{\textbf{Advanced Cyberinfrastructure Coordination Ecosystem: Services \& Support}} (ACCESS) program, which is supported by U.S. National Science Foundation grants \#2138259, \#2138286, \#2138307, \#2137603, and \#2138296.

\bibliography{iclr2027_conference}
\bibliographystyle{iclr2027_conference}

\appendix

\section{Prompts for synthetic task evaluation}
\label{app:synthetic_evaluation}

As discussed in section~\ref{sec:experiment}, for each number of features $d$, we construct $49 \times 5$ $D_{\mathrm{obs}}$, corresponding to 49 true probabilities and five random shuffles per probability. We provide each $D_{\mathrm{obs}}$ to the model as in-context signals, and then prompt each LLM separately for beliefs and decisions. 

We elicit the model's belief with a proper scoring rule five times, using the following template. After obtaining LLMs' responses, we average the five beliefs for the same true probability (but different shuffles) as the model's reported belief. This reduces bias due to the model reacting to the order of the provided samples.

\begin{notebox}
You are given observed data from previous samples. Each row is one previous sample with binary features and a binary outcome. Feature values are 0 or 1. The column outcome is 1 for positive outcome and 0 for negative outcome. Use only the observed data to estimate the probability that a new sample with the specified feature values has outcome = 1.

\vspace{0.8em}

Observed data:
\{$D_{obs}$\}

\vspace{0.8em}

New sample feature values:
\{feature vector\}

\vspace{0.8em}

Your reported probability p (between 0 and 1) will be evaluated against the true outcome by using a quadratic scoring rule:

If the event is TRUE: Score = $1 - (1 - p)^2$

If the event is FALSE: Score = $1 - p^2$

Under the scoring rule, the highest expected score comes from reporting your true subjective probability exactly. Based on the observed data, what is your true subjective probability that the new sample with the specified feature values has outcome = 1?

\vspace{0.8em}

Respond with EXACTLY one line, and no extra text. The line MUST be in the format: 'Probability: \textless number between 0 and 1\textgreater'

Example:

Probability: 0.37
\end{notebox}

We ask for the model's decision under the 9 thresholds. For threshold decision prompts, we use:

\begin{notebox}
You are given observed data from previous samples. Each row is one previous sample with binary features and a binary outcome. Feature values are 0 or 1. The column outcome is 1 for positive outcome and 0 for negative outcome. Use only the observed data to decide whether to assign the positive label to a new sample with the specified feature values under the given cost table.

\vspace{0.8em}

Observed data:
\{$D_{obs}$\}

\vspace{0.8em}

New sample feature values:
\{feature vector\}

\vspace{0.8em}

Decision task:

Question: Will you assign the positive label to this new sample given the costs below?

\vspace{0.8em}

Costs:

        - If you assign the positive label and the true outcome is negative (false positive), cost = \{$C_{FP}$\}
        
        - If you assign the negative label and the true outcome is positive (false negative), cost = \{$C_{FN}$\}
        
        - Correct decisions have cost = 0
        
Choose the label with lower expected cost.

\vspace{0.8em}

Respond with EXACTLY one line, and no extra text. The line MUST be in the format: 'Decision: \textless 0 or 1\textgreater'

Example:

Decision: 0
\end{notebox}

\section{Synthetic task}
\label{app:synthetic}

The post-training experiments use the same data-generating model as in Section~\ref{sec:diagnosis}. We independently generate 1000 evidence cases with five binary features, each contains a observed dataset of 300 examples and a test case. We randomly split 700 evidence cases for training, 100 for validation, and 200 for final evaluation.

\textbf{Belief prompt}

\begin{notebox}
You are given observed data from previous samples. Each row is one previous sample with binary features and a binary outcome.
Feature values are 0 or 1. The column y is 1 for positive outcome and 0 for negative outcome.

Observed data:
\{observed\_data\}

New sample feature values:
\{test\_case\}

Estimate the probability that the new sample has outcome y = 1.

You may reason concisely. Then output a final answer line exactly in the format:
Probability: 0.xxx
\end{notebox}

\textbf{Sequential decision prompt}

\begin{notebox}
Now make the decision: 
    
Will you assign the positive label to this new sample given the costs below?
    
Choose action 1 to assign the positive label.

Choose action 0 to assign the negative label.

Costs:

- If you choose action 1 and the true outcome is 0, false positive cost = \{c\_fp\}

- If you choose action 0 and the true outcome is 1, false negative cost = \{c\_fn\}

- Correct decisions have cost = 0

You may reason concisely. Then output a final answer line exactly in the format:
Decision: 0 or 1
\end{notebox}

\textbf{Direct decision prompt}

\begin{notebox}
You are given observed data from previous samples. Each row is one previous sample with binary features and a binary outcome.
Feature values are 0 or 1. The column y is 1 for positive outcome and 0 for negative outcome.

Observed data:
\{observed\_data\}

New sample feature values:
\{test\_case\}

Now make the decision: 
    
Will you assign the positive label to this new sample given the costs below?
    
Choose action 1 to assign the positive label.

Choose action 0 to assign the negative label.

Costs:

- If you choose action 1 and the true outcome is 0, false positive cost = \{c\_fp\}

- If you choose action 0 and the true outcome is 1, false negative cost = \{c\_fn\}

- Correct decisions have cost = 0

You may reason concisely. Then output a final answer line exactly in the format:
Decision: 0 or 1
\end{notebox}

\section{Weather forecasting task}
\label{app:hail}

We construct this task from the HailFinder Bayesian network~\citep{abramson1996hailfinder}. Each evidence contains seven upstream atmospheric variables: \texttt{AMInstabMt}, \texttt{CldShadeOth}, \texttt{LatestCIN}, \texttt{LLIW}, \texttt{ScnRelPlFcst}, \texttt{InsSclInScen}, and \texttt{CapInScen}. We render their values as short descriptions grouped into mountain and plains conditions. The binary outcome indicates whether the Region 5 forecast is significant or severe hail (\texttt{R5Fcst} $\in{\texttt{SIG},\texttt{SVR}}$). For each distinct observed context $e$, we compute the exact reference probability $p^*(e)=P(Y=1\mid e)$ from the network. We split contexts, with no overlap, into 3,000 training, 300 validation, and 500 evaluation examples.

\textbf{Belief prompt}

\begin{notebox}
You are forecasting the hail category for the entire Denver warning area.

Weather observations:
\{evidence\_text\}

Estimate the probability that the hail category for the entire Denver warning area will be significant or severe.

You may reason concisely. Then output a final answer line exactly in the format:
Probability: 0.xxx
\end{notebox}

\textbf{Sequential decision prompt}

\begin{notebox}
    Now choose a severe-weather warning action under these losses:
    
- Action 0: do not issue a significant-or-severe hail warning. Loss = \{c\_fn\} if the true weather is significant or severe hail, otherwise 0.

- Action 1: issue a significant-or-severe hail warning. Loss = \{c\_fp\} if the true weather is no significant hail, otherwise 0.

You may reason concisely. Then output a final answer line exactly in the format:
Decision: 0 or 1
\end{notebox}

\textbf{Direct decision prompt}

\begin{notebox}
  You are forecasting the hail category for the entire Denver warning area.

Weather observations:
\{evidence\_text\}

Now choose a severe-weather warning action under these losses:
    
- Action 0: do not issue a significant-or-severe hail warning. Loss = \{c\_fn\} if the true weather is significant or severe hail, otherwise 0.

- Action 1: issue a significant-or-severe hail warning. Loss = \{c\_fp\} if the true weather is no significant hail, otherwise 0.

You may reason concisely. Then output a final answer line exactly in the format:
Decision: 0 or 1
\end{notebox}

\section{Clinical task}
\label{app:simsum}

We use SimSUM's~\citep{rabaey2025simsum} 10,000 synthetic patient records, which pair clinical notes with variables sampled from a known Bayesian network. The outcome is the presence of pneumonia or a common cold. We filter records with high-concerned symptoms. This includes dyspnea, high fever, or both cough and pain. The model receives the clinical note; while we compute posteriors from the network conditional on the corresponding recorded health conditions and symptoms, marginalizing unobserved variables. We split all the retained cases into 2,211 training, 300 validation, and 500 evaluation examples, stratified by outcome and reference probability.

\textbf{Belief prompt}

\begin{notebox}
You're assessing whether a patient with higher-concern symptoms has an infectious respiratory condition.

Clinical note:
\{clinical\_note\}

Estimate the probability that this patient has an infectious respiratory condition (pneumonia or the common cold).

You may reason concisely. Then output a final answer line exactly in the format:
Probability: 0.xxx
\end{notebox}

\textbf{Sequential decision prompt}

\begin{notebox}
Now choose a clinical action under these losses:

- Action 0: no infection-focused escalation. Loss = \{c\_fn\} if an infectious respiratory condition is present, otherwise 0.

- Action 1: escalate for additional infection-focused clinical evaluation. Loss = \{c\_fp\} if an infectious respiratory condition is absent, otherwise 0.

You may reason concisely. Then output a final answer line exactly in the format:
Decision: 0 or 1
\end{notebox}

\textbf{Direct decision prompt}

\begin{notebox}
You're assessing whether a patient with higher-concern symptoms has an infectious respiratory condition.

Clinical note:
\{clinical\_note\}

Now choose a clinical action under these losses:

- Action 0: no infection-focused escalation. Loss = \{c\_fn\} if an infectious respiratory condition is present, otherwise 0.

- Action 1: escalate for additional infection-focused clinical evaluation. Loss = \{c\_fp\} if an infectious respiratory condition is absent, otherwise 0.

You may reason concisely. Then output a final answer line exactly in the format:
Decision: 0 or 1
\end{notebox}

%\clearpage

\section{Full intervention results}
\label{app:results}

% Requires \usepackage{booktabs}

\begin{table}[ht]
  \centering
  \small
    \begin{tabular}{llccccc}
    \toprule
    \textbf{Strategy} & \textbf{Format} & $\Delta R_{\mathrm{total}}$ & $\Delta R_{\mathrm{belief}}$ & $\Delta R_{\mathrm{opt}}$ & $\Delta \mathrm{Belief\_MSE}$ & $\Delta \mathrm{Agreement}$ \\
    \midrule
    B & independent & $-0.031 \pm 0.010$ & $-0.045 \pm 0.012$ & $0.014 \pm 0.014$ & $-0.072 \pm 0.017$ & $0.095 \pm 0.036$ \\
    B & two-turn & $-0.046 \pm 0.011$ & $-0.046 \pm 0.012$ & $0.000 \pm 0.002$ & $-0.072 \pm 0.017$ & $-0.001 \pm 0.003$ \\
    D & independent & $-0.015 \pm 0.010$ & $-0.013 \pm 0.012$ & $-0.002 \pm 0.015$ & $-0.015 \pm 0.017$ & $0.063 \pm 0.034$ \\
    D & two-turn & $-0.010 \pm 0.012$ & $-0.012 \pm 0.012$ & $0.001 \pm 0.002$ & $-0.015 \pm 0.017$ & $0.001 \pm 0.002$ \\
    B/D & independent & $-0.042 \pm 0.010$ & $-0.048 \pm 0.012$ & $0.006 \pm 0.015$ & $-0.071 \pm 0.017$ & $0.170 \pm 0.034$ \\
    B/D & two-turn & $-0.042 \pm 0.011$ & $-0.046 \pm 0.011$ & $0.003 \pm 0.003$ & $-0.071 \pm 0.017$ & $-0.015 \pm 0.006$ \\
    B+A & independent & $-0.041 \pm 0.011$ & $-0.046 \pm 0.012$ & $0.005 \pm 0.016$ & $-0.071 \pm 0.017$ & $0.156 \pm 0.037$ \\
    B+A & two-turn & $-0.046 \pm 0.011$ & $-0.046 \pm 0.011$ & $0.001 \pm 0.002$ & $-0.072 \pm 0.017$ & $-0.002 \pm 0.004$ \\
    \bottomrule
    \end{tabular}
  \caption{Synthetic. Changes from the Qwen3-8B baseline. Mean difference with 95\% confidence interval (n = 200 cases). Negative values indicate lower regret than Base.}
  \label{tab:r_syn}
\end{table}

\begin{table}[ht]
  \centering
  \small
    \begin{tabular}{llccccc}
    \toprule
    \textbf{Strategy} & \textbf{Format} & $\Delta R_{\mathrm{total}}$ & $\Delta R_{\mathrm{belief}}$ & $\Delta R_{\mathrm{opt}}$ & $\Delta \mathrm{Belief\_MSE}$ & $\Delta \mathrm{Agreement}$ \\
    \midrule
    B & independent & $-0.024 \pm 0.010$ & $-0.125 \pm 0.014$ & $0.101 \pm 0.016$ & $-0.130 \pm 0.017$ & $-0.164 \pm 0.040$ \\
    B & two-turn & $-0.111 \pm 0.015$ & $-0.125 \pm 0.014$ & $0.014 \pm 0.006$ & $-0.130 \pm 0.016$ & $-0.039 \pm 0.014$ \\
    D & independent & $-0.184 \pm 0.019$ & $-0.020 \pm 0.006$ & $-0.165 \pm 0.019$ & $-0.024 \pm 0.006$ & $-0.119 \pm 0.021$ \\
    D & two-turn & $-0.032 \pm 0.007$ & $-0.020 \pm 0.006$ & $-0.013 \pm 0.004$ & $-0.024 \pm 0.006$ & $-0.003 \pm 0.007$ \\
    B/D & independent & $-0.159 \pm 0.019$ & $-0.136 \pm 0.017$ & $-0.023 \pm 0.018$ & $-0.137 \pm 0.021$ & $-0.002 \pm 0.045$ \\
    B/D & two-turn & $-0.123 \pm 0.015$ & $-0.136 \pm 0.017$ & $0.012 \pm 0.009$ & $-0.137 \pm 0.021$ & $-0.120 \pm 0.020$ \\
    B+A & independent & $-0.008 \pm 0.006$ & $-0.136 \pm 0.017$ & $0.128 \pm 0.018$ & $-0.137 \pm 0.021$ & $-0.329 \pm 0.044$ \\
    B+A & two-turn & $-0.122 \pm 0.016$ & $-0.136 \pm 0.017$ & $0.013 \pm 0.008$ & $-0.137 \pm 0.021$ & $-0.089 \pm 0.019$ \\
    \bottomrule
    \end{tabular}
  \caption{Weather. Changes from the Qwen3-8B baseline. Mean difference with 95\% confidence interval (n = 500 cases). Negative values indicate lower regret than Base.}
  \label{tab:r_hail}
\end{table}

\begin{table}[htb]
  \centering
  \small
    \begin{tabular}{llccccc}
    \toprule
    \textbf{Strategy} & \textbf{Format} & $\Delta R_{\mathrm{total}}$ & $\Delta R_{\mathrm{belief}}$ & $\Delta R_{\mathrm{opt}}$ & $\Delta \mathrm{Belief\_MSE}$ & $\Delta \mathrm{Agreement}$ \\
    \midrule
    B & independent & $-0.025 \pm 0.017$ & $-0.079 \pm 0.013$ & $0.054 \pm 0.022$ & $-0.089 \pm 0.015$ & $-0.038 \pm 0.032$ \\
    B & two-turn & $-0.037 \pm 0.015$ & $-0.079 \pm 0.013$ & $0.042 \pm 0.016$ & $-0.089 \pm 0.015$ & $-0.105 \pm 0.024$ \\
    D & independent & $-0.180 \pm 0.023$ & $-0.035 \pm 0.010$ & $-0.145 \pm 0.023$ & $-0.040 \pm 0.011$ & $0.177 \pm 0.028$ \\
    D & two-turn & $-0.073 \pm 0.013$ & $-0.035 \pm 0.010$ & $-0.038 \pm 0.010$ & $-0.040 \pm 0.011$ & $0.049 \pm 0.014$ \\
    B/D & independent & $-0.153 \pm 0.023$ & $-0.069 \pm 0.013$ & $-0.084 \pm 0.025$ & $-0.079 \pm 0.015$ & $0.155 \pm 0.035$ \\
    B/D & two-turn & $-0.089 \pm 0.014$ & $-0.069 \pm 0.013$ & $-0.020 \pm 0.010$ & $-0.079 \pm 0.015$ & $0.028 \pm 0.015$ \\
    B+A & independent & $-0.042 \pm 0.014$ & $-0.076 \pm 0.012$ & $0.034 \pm 0.019$ & $-0.086 \pm 0.014$ & $-0.025 \pm 0.030$ \\
    B+A & two-turn & $-0.114 \pm 0.015$ & $-0.076 \pm 0.012$ & $-0.039 \pm 0.010$ & $-0.086 \pm 0.014$ & $0.090 \pm 0.013$ \\
    \bottomrule
    \end{tabular}
  \caption{Clinical. Changes from the Qwen3-8B baseline. Mean difference with 95\% confidence interval (n = 500 cases). Negative values indicate lower regret than Base.}
  \label{tab:}
\end{table}

\clearpage %

\section{Full diagnosis results}
\label{app:synthetic_results}

\begin{figure}[htb]
    \centering
    \includegraphics[width=0.95\textwidth]{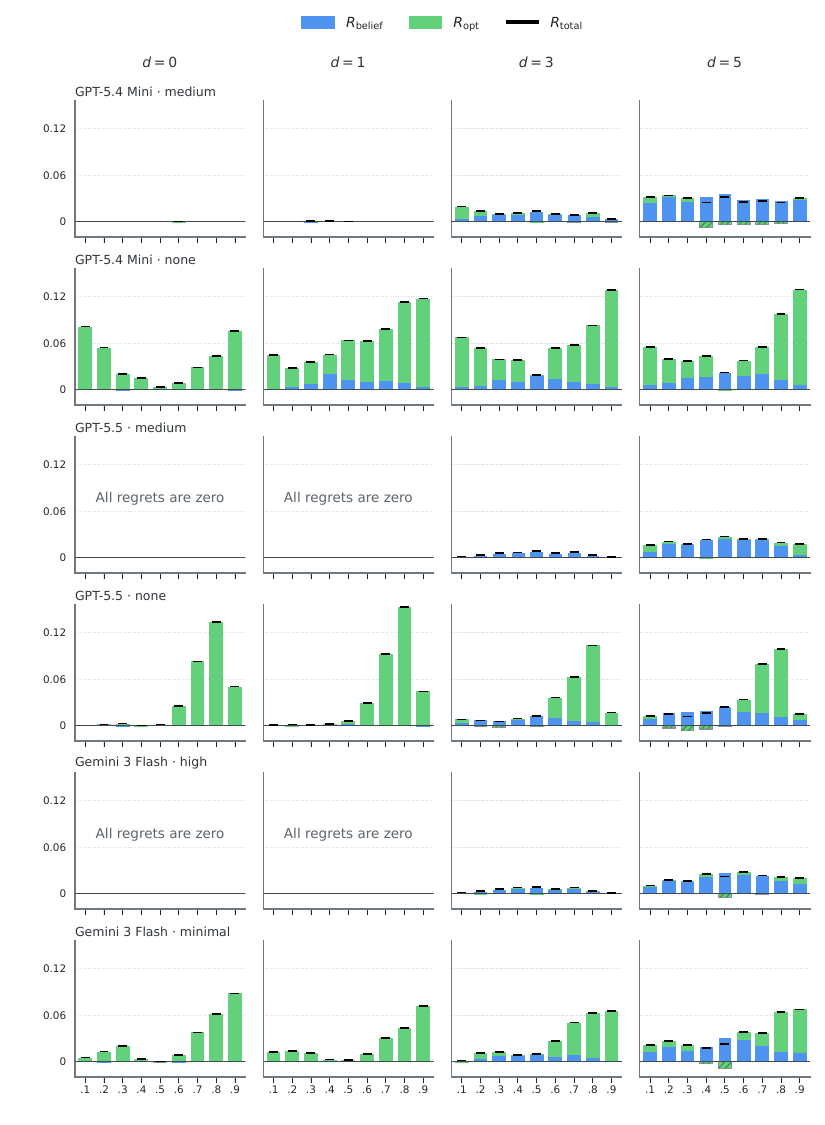}
    \caption{Full normalized regret decomposition on the synthetic task across all model--reasoning settings. Columns vary feature size $d \in \{0,1,3,5\}$; the regret decomposition follows Figure~\ref{fig:main-decomposition}. ``All regrets are zero'' marks settings that are optimal at every threshold.}
    \label{fig:full-1}
\end{figure}

\begin{figure}[htb]
    \centering
    \ContinuedFloat
    \includegraphics[width=0.95\textwidth]{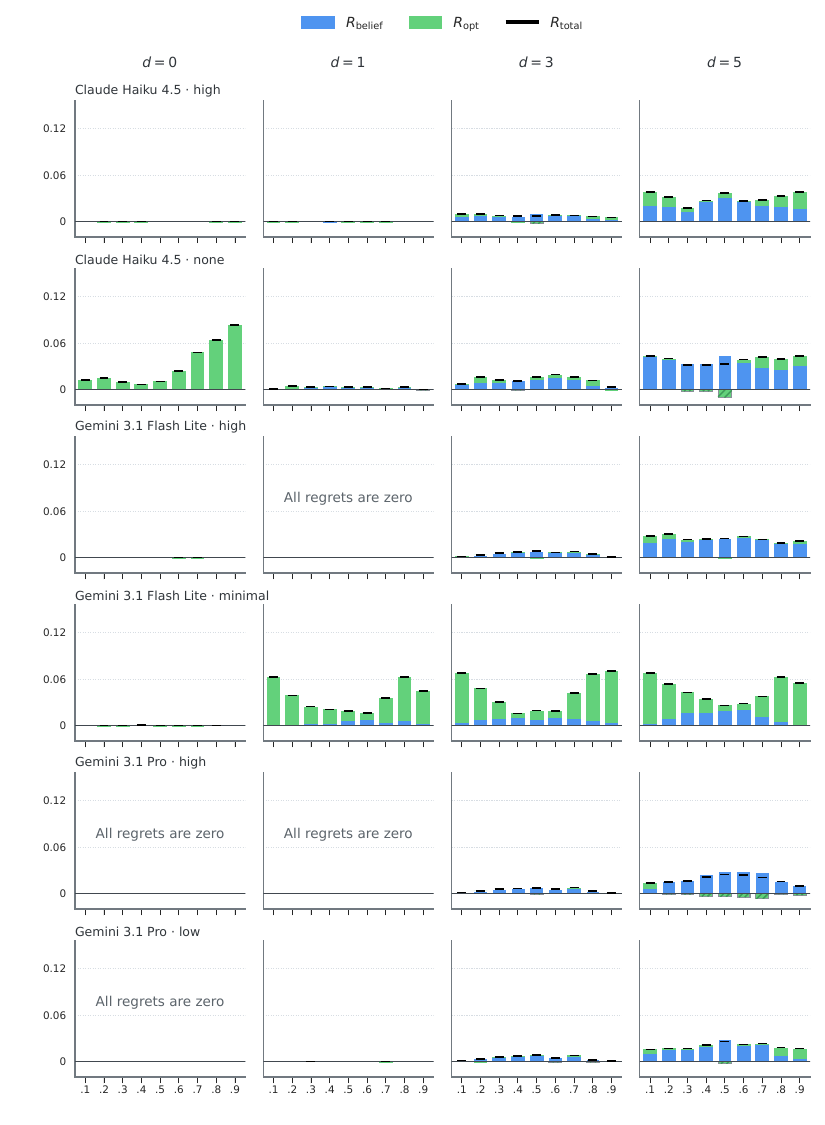}
    \caption[]{Figure ~\ref{fig:full-1} continued.}
    \label{fig:full-2}
\end{figure}

\begin{figure}[htb]
    \centering
    \ContinuedFloat
    \includegraphics[width=0.95\textwidth]{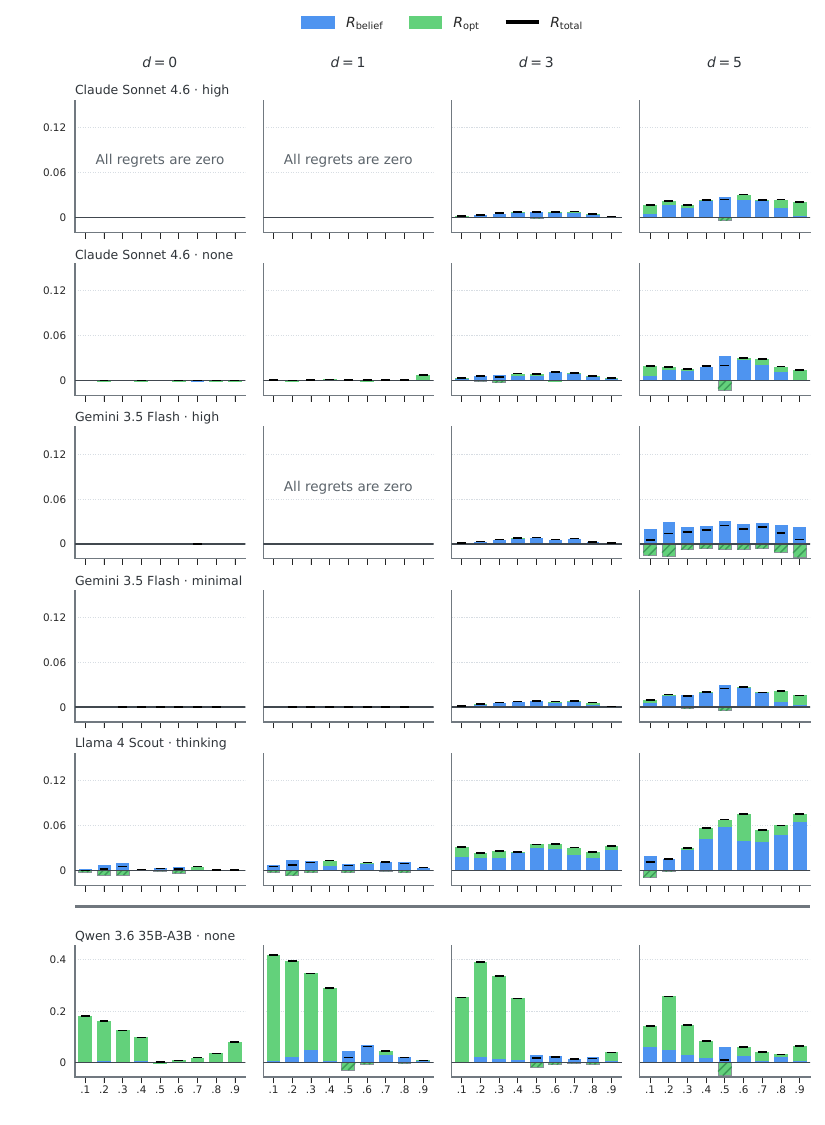}
    \caption[]{Figure ~\ref{fig:full-1} continued. Note the different y-axis scaling for Qwen 3.6 35B-A3B, which has considerably higher regret.}
    \label{fig:full-3}
\end{figure}

\begin{table}[t]
\centering
\caption{Regret under the true posterior ($d=3$).}
\label{tab:true-posterior-regret-d3}
\scriptsize
\setlength{\tabcolsep}{1pt}
\begin{tabular*}{\linewidth}{@{\extracolsep{\fill}}lrrr@{}}
\hline
Model / reasoning mode & $\bar{R}_{\mathrm{total}}$ & $\bar{R}_{\mathrm{belief}}$ & $\bar{R}_{\mathrm{opt}}$ \\
\hline
Claude Haiku 4.5 / high & $0.008\!\pm\!0.002$ & $0.007\!\pm\!0.003$ & $0.001\!\pm\!0.003$ \\
Claude Haiku 4.5 / none & $0.013\!\pm\!0.003$ & $0.009\!\pm\!0.003$ & $0.003\!\pm\!0.003$ \\
Claude Sonnet 4.6 / high & $0.005\!\pm\!0.002$ & $0.005\!\pm\!0.002$ & $0.000\!\pm\!0.001$ \\
Claude Sonnet 4.6 / none & $0.007\!\pm\!0.003$ & $0.006\!\pm\!0.002$ & $0.001\!\pm\!0.002$ \\
Gemini 3.1 Flash Lite / high & $0.005\!\pm\!0.002$ & $0.005\!\pm\!0.002$ & $0.000\!\pm\!0.001$ \\
Gemini 3.1 Flash Lite / minimal & $0.042\!\pm\!0.005$ & $0.007\!\pm\!0.002$ & $0.035\!\pm\!0.005$ \\
Gemini 3.1 Pro / high & $0.005\!\pm\!0.002$ & $0.005\!\pm\!0.002$ & $0.000\!\pm\!0.001$ \\
Gemini 3.1 Pro / low & $0.005\!\pm\!0.002$ & $0.005\!\pm\!0.002$ & $0.000\!\pm\!0.001$ \\
Gemini 3.5 Flash / high & $0.005\!\pm\!0.002$ & $0.005\!\pm\!0.002$ & $0.000\!\pm\!0.001$ \\
Gemini 3.5 Flash / minimal & $0.006\!\pm\!0.002$ & $0.005\!\pm\!0.002$ & $0.001\!\pm\!0.002$ \\
Gemini 3 Flash / high & $0.005\!\pm\!0.002$ & $0.005\!\pm\!0.002$ & $0.000\!\pm\!0.001$ \\
Gemini 3 Flash / minimal & $0.028\!\pm\!0.005$ & $0.005\!\pm\!0.002$ & $0.023\!\pm\!0.005$ \\
GPT-5.4 Mini / medium & $0.011\!\pm\!0.003$ & $0.008\!\pm\!0.003$ & $0.003\!\pm\!0.004$ \\
GPT-5.4 Mini / none & $0.060\!\pm\!0.005$ & $0.009\!\pm\!0.002$ & $0.051\!\pm\!0.006$ \\
GPT-5.5 / medium & $0.005\!\pm\!0.002$ & $0.005\!\pm\!0.002$ & $0.000\!\pm\!0.001$ \\
GPT-5.5 / none & $0.029\!\pm\!0.004$ & $0.007\!\pm\!0.002$ & $0.022\!\pm\!0.005$ \\
Llama 4 Scout / thinking & $0.029\!\pm\!0.006$ & $0.022\!\pm\!0.005$ & $0.007\!\pm\!0.005$ \\
Qwen 3.6 35B-A3B / none & $0.149\!\pm\!0.006$ & $0.017\!\pm\!0.004$ & $0.131\!\pm\!0.006$ \\
\hline
\end{tabular*}
\par\smallskip
\parbox{\linewidth}{\footnotesize 
Regrets are normalized and averaged over nine decision thresholds.
Entries show the estimate $\pm$ the larger distance to either endpoint of the
95\% stratified paired bootstrap interval, rounded outward.}
\end{table}

\begin{table}[t]
\centering
\caption{Regret under the true posterior ($d=5$).}
\label{tab:true-posterior-regret-d5}
\scriptsize
\setlength{\tabcolsep}{1pt}
\begin{tabular*}{\linewidth}{@{\extracolsep{\fill}}lrrr@{}}
\hline
Model / reasoning mode & $\bar{R}_{\mathrm{total}}$ & $\bar{R}_{\mathrm{belief}}$ & $\bar{R}_{\mathrm{opt}}$ \\
\hline
Claude Haiku 4.5 / high & $0.031\!\pm\!0.005$ & $0.021\!\pm\!0.005$ & $0.010\!\pm\!0.004$ \\
Claude Haiku 4.5 / none & $0.038\!\pm\!0.006$ & $0.034\!\pm\!0.007$ & $0.004\!\pm\!0.007$ \\
Claude Sonnet 4.6 / high & $0.022\!\pm\!0.005$ & $0.016\!\pm\!0.004$ & $0.007\!\pm\!0.003$ \\
Claude Sonnet 4.6 / none & $0.020\!\pm\!0.005$ & $0.016\!\pm\!0.004$ & $0.004\!\pm\!0.003$ \\
Gemini 3.1 Flash Lite / high & $0.024\!\pm\!0.005$ & $0.022\!\pm\!0.005$ & $0.002\!\pm\!0.004$ \\
Gemini 3.1 Flash Lite / minimal & $0.045\!\pm\!0.006$ & $0.011\!\pm\!0.003$ & $0.034\!\pm\!0.006$ \\
Gemini 3.1 Pro / high & $0.018\!\pm\!0.004$ & $0.020\!\pm\!0.007$ & $-0.002\!\pm\!0.006$ \\
Gemini 3.1 Pro / low & $0.020\!\pm\!0.004$ & $0.016\!\pm\!0.004$ & $0.004\!\pm\!0.003$ \\
Gemini 3.5 Flash / high & $0.016\!\pm\!0.004$ & $0.026\!\pm\!0.007$ & $-0.010\!\pm\!0.007$ \\
Gemini 3.5 Flash / minimal & $0.019\!\pm\!0.004$ & $0.016\!\pm\!0.004$ & $0.003\!\pm\!0.003$ \\
Gemini 3 Flash / high & $0.020\!\pm\!0.005$ & $0.018\!\pm\!0.004$ & $0.002\!\pm\!0.003$ \\
Gemini 3 Flash / minimal & $0.035\!\pm\!0.005$ & $0.019\!\pm\!0.004$ & $0.016\!\pm\!0.005$ \\
GPT-5.4 Mini / medium & $0.029\!\pm\!0.004$ & $0.029\!\pm\!0.006$ & $0.000\!\pm\!0.005$ \\
GPT-5.4 Mini / none & $0.057\!\pm\!0.006$ & $0.014\!\pm\!0.003$ & $0.043\!\pm\!0.006$ \\
GPT-5.5 / medium & $0.021\!\pm\!0.005$ & $0.017\!\pm\!0.004$ & $0.005\!\pm\!0.003$ \\
GPT-5.5 / none & $0.034\!\pm\!0.004$ & $0.015\!\pm\!0.004$ & $0.019\!\pm\!0.005$ \\
Llama 4 Scout / thinking & $0.049\!\pm\!0.006$ & $0.039\!\pm\!0.008$ & $0.010\!\pm\!0.008$ \\
Qwen 3.6 35B-A3B / none & $0.093\!\pm\!0.007$ & $0.031\!\pm\!0.006$ & $0.061\!\pm\!0.008$ \\
\hline
\end{tabular*}
\par\smallskip
\parbox{\linewidth}{\footnotesize
Regrets are normalized and averaged equally over nine decision thresholds.
Entries show the estimate $\pm$ the larger distance to either endpoint of the
95\% stratified paired bootstrap interval, rounded outward.}
\end{table}

\clearpage %

\section{Elicited vs. revealed beliefs}
\label{app:revealed_beliefs}

Schervish's representation theorem implies that proper scoring rules for binary beliefs can be expressed as mixtures of elementary threshold losses. The quadratic score satisfies
\[
(p-y)^2 = 2\int_0^1 \ell_{\tau}(\mathbf{1}\{p\ge \tau\},y)\,d\tau.
\]

If beliefs are linear, then optimal actions satisfy:

\[
a_{\tau}^*(z)=\mathbf{1}\{\beta_0+\beta^\top z \ge \tau\}.
\]

We use this construction to test the extent to which belief representations obtained through prompting with a proper scoring rule versus inferred from a model's repeated decisions appear equivalent.
For instances in which the model’s threshold decisions exhibit a monotonic sequence, we compare the reported belief with the decision-revealed belief inferred from the unique switch point. Table~\ref{tab:monotonic_rate} shows the monotonic rates across feature sizes and models.

\begin{table}[ht]
  \centering

\begin{tabular}{lcccc}
\hline
\multicolumn{1}{c}{}             & \multicolumn{1}{c}{\textbf{Zero feature}} & \multicolumn{1}{c}{\textbf{One feature}} & \multicolumn{1}{c}{\textbf{Three features}} & \multicolumn{1}{c}{\textbf{Five features}} \\ \hline
\textbf{Claude-Sonnet-4.6\_High} & 1.000                                     & 1.000                                    & 1.000                                       & 0.955                                      \\
\textbf{Gemini-3.5-Flash\_High}  & 1.000                                     & 1.000                                    & 1.000                                       & 0.984                                      \\
\textbf{GPT-5.5\_Medium}         & 1.000                                     & 1.000                                    & 1.000                                       & 1.000                                      \\
\textbf{GPT-5.5\_None}           & 0.853                                     & 0.820                                    & 0.902                                       & 0.841                                      \\
\textbf{Llama-4\_Thinking}       & 1.000                                     & 0.939                                    & 0.878                                       & 0.563                                      \\
\textbf{Qwen-3.6\_None}          & 0.714                                     & 0.588                                    & 0.453                                       & 0.445                                      \\ \hline
\end{tabular}
  \caption{Monotonic rates across feature sizes and models}
  \label{tab:monotonic_rate}
\end{table}

\begin{figure}[htb]
    \centering
    \includegraphics[width=0.95\textwidth]{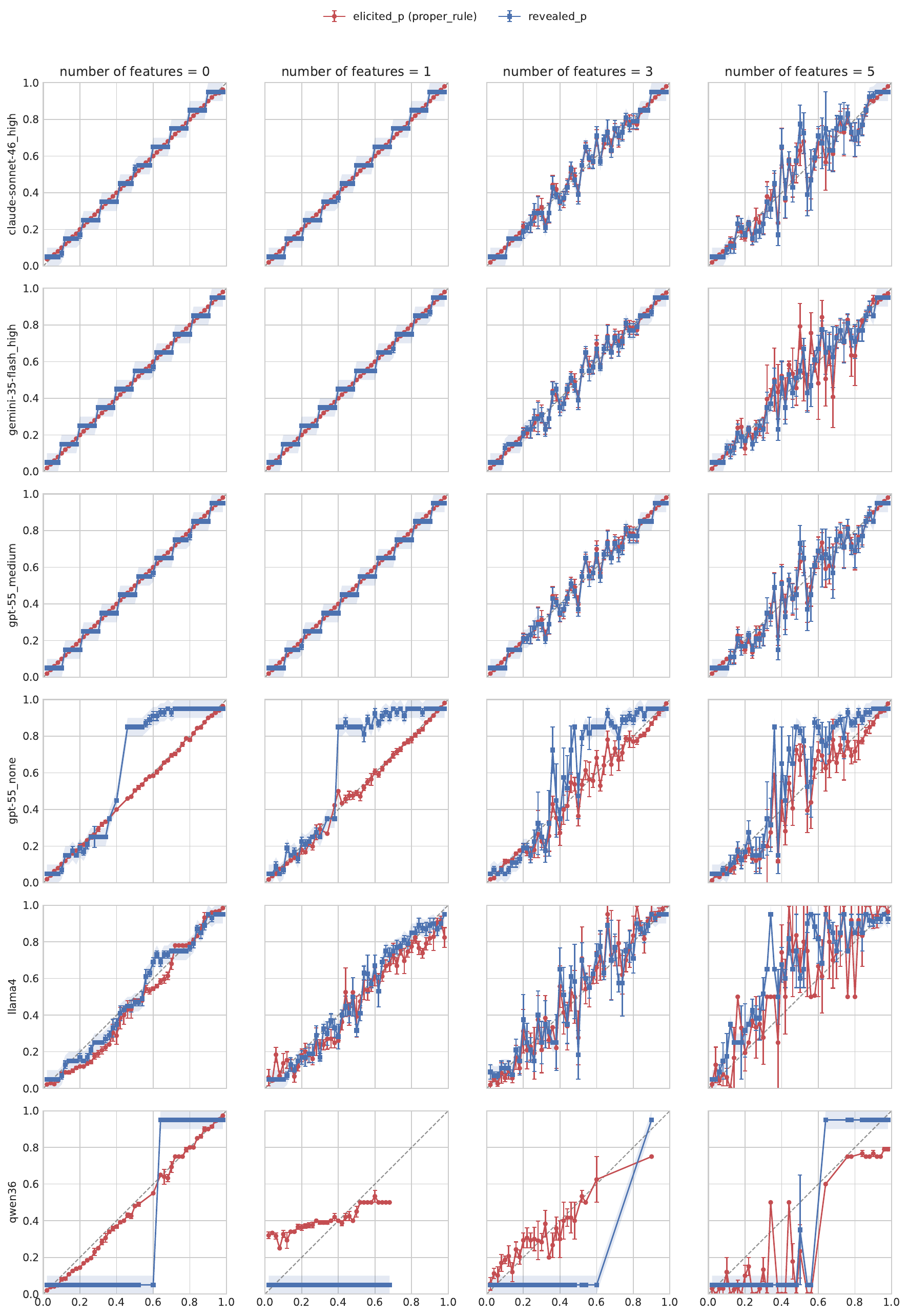}
    \caption{Comparisons between elicited and decision-revealed beliefs by feature size and model}
    \label{fig:belief_comparison}
\end{figure}

Figure~\ref{fig:belief_comparison} shows the comparison between reported and decision-revealed beliefs. Among the selected model settings, Claude Sonnet~4.6 (high), Gemini~3.5 Flash (high), and GPT-5.5 (medium) exhibit smoother and more closely aligned elicited and decision-revealed beliefs than Llama~4 with thinking and non-thinking Qwen~3.6. Within GPT-5.5, the setting with no thinking shows greater misalignment than medium thinking, including at small feature sizes. For most settings, both belief measures fluctuate more as feature size increases.

\section{True posterior vs. empirical frequency as reference}
\label{app:posterior_empirical_frequency}

The regret decomposition on the synthetic task in section \ref{sec:diagnosis} uses the true posterior, which depends on the number of active features in the test vector. As an alternative reference, we can also use the empirical frequency, which is the share of positive outcomes among rows whose full feature vector matches the test vector. The two references coincide at feature sizes $d=0$ and $d=1$. However, with three or five features, the true posterior pools different vectors with the same number of active features, whereas the empirical frequency uses only exact matches. At feature sizes $d=3$ and $d=5$, we recompute the decomposition on the same observations, changing only the reference from the true posterior to the empirical frequency. Results are in the following tables.

After averaging over model settings and decision thresholds, mean normalized total regret decreases by 18\% at $d=3$ and 25\% at $d=5$. This decrease suggests that, with our evaluation results, model decisions are better aligned with empirical frequencies than with the true posterior.  At $d=3$, the decrease is primarily in belief loss. At $d=5$, both belief loss and the signed optimization residual decrease on average, especially GPT-5.5 with medium thinking, which shows a particularly large decrease in both terms at $d=5$. Also, there are two exceptions. For Qwen 3.6 35B-A3B without thinking, the increase in belief loss is larger than the decrease in optimization residual, resulting in higher total regret; for Llama 4 Scout, both belief loss and optimization residual increase.

\begin{table}[t]
\centering
\caption{Changes in regret and stated-belief error under empirical frequency versus the true posterior ($d=3$).}
\label{tab:posterior-frequency-comparison-d3}
\scriptsize
\setlength{\tabcolsep}{1pt}
\begin{tabular*}{\linewidth}{@{\extracolsep{\fill}}lrrrr@{}}
\hline
Model / reasoning mode & $\Delta R_{\mathrm{total}}$ & $\Delta R_{\mathrm{belief}}$ & $\Delta R_{\mathrm{opt}}$ & $\Delta\mathrm{MSE}$ \\
\hline
Claude Haiku 4.5 / high & $-0.005\!\pm\!0.002$ & $-0.006\!\pm\!0.002$ & $0.000\!\pm\!0.002$ & $-0.007\!\pm\!0.002$ \\
Claude Haiku 4.5 / none & $-0.004\!\pm\!0.003$ & $-0.003\!\pm\!0.003$ & $-0.001\!\pm\!0.002$ & $-0.004\!\pm\!0.003$ \\
Claude Sonnet 4.6 / high & $-0.005\!\pm\!0.002$ & $-0.005\!\pm\!0.002$ & $0.000\!\pm\!0.001$ & $-0.006\!\pm\!0.002$ \\
Claude Sonnet 4.6 / none & $-0.004\!\pm\!0.002$ & $-0.004\!\pm\!0.002$ & $0.000\!\pm\!0.002$ & $-0.005\!\pm\!0.002$ \\
Gemini 3.1 Flash Lite / high & $-0.005\!\pm\!0.002$ & $-0.005\!\pm\!0.002$ & $0.000\!\pm\!0.001$ & $-0.006\!\pm\!0.002$ \\
Gemini 3.1 Flash Lite / minimal & $-0.005\!\pm\!0.004$ & $-0.001\!\pm\!0.002$ & $-0.004\!\pm\!0.004$ & $-0.002\!\pm\!0.002$ \\
Gemini 3.1 Pro / high & $-0.005\!\pm\!0.002$ & $-0.005\!\pm\!0.002$ & $0.000\!\pm\!0.001$ & $-0.006\!\pm\!0.002$ \\
Gemini 3.1 Pro / low & $-0.005\!\pm\!0.002$ & $-0.005\!\pm\!0.002$ & $0.000\!\pm\!0.001$ & $-0.006\!\pm\!0.002$ \\
Gemini 3.5 Flash / high & $-0.005\!\pm\!0.002$ & $-0.005\!\pm\!0.002$ & $0.000\!\pm\!0.001$ & $-0.006\!\pm\!0.002$ \\
Gemini 3.5 Flash / minimal & $-0.005\!\pm\!0.002$ & $-0.005\!\pm\!0.002$ & $0.000\!\pm\!0.001$ & $-0.006\!\pm\!0.002$ \\
Gemini 3 Flash / high & $-0.005\!\pm\!0.002$ & $-0.005\!\pm\!0.002$ & $0.000\!\pm\!0.001$ & $-0.006\!\pm\!0.002$ \\
Gemini 3 Flash / minimal & $-0.007\!\pm\!0.003$ & $-0.004\!\pm\!0.002$ & $-0.004\!\pm\!0.003$ & $-0.005\!\pm\!0.002$ \\
GPT-5.4 Mini / medium & $-0.004\!\pm\!0.002$ & $-0.005\!\pm\!0.003$ & $+0.001\!\pm\!0.002$ & $-0.006\!\pm\!0.003$ \\
GPT-5.4 Mini / none & $0.000\!\pm\!0.003$ & $-0.002\!\pm\!0.002$ & $+0.003\!\pm\!0.003$ & $-0.002\!\pm\!0.002$ \\
GPT-5.5 / medium & $-0.005\!\pm\!0.002$ & $-0.005\!\pm\!0.002$ & $0.000\!\pm\!0.001$ & $-0.006\!\pm\!0.002$ \\
GPT-5.5 / none & $-0.008\!\pm\!0.003$ & $-0.005\!\pm\!0.002$ & $-0.003\!\pm\!0.003$ & $-0.006\!\pm\!0.002$ \\
Llama 4 Scout / thinking & $-0.002\!\pm\!0.004$ & $-0.001\!\pm\!0.004$ & $-0.002\!\pm\!0.003$ & $-0.002\!\pm\!0.005$ \\
Qwen 3.6 35B-A3B / none & $+0.003\!\pm\!0.005$ & $0.000\!\pm\!0.003$ & $+0.003\!\pm\!0.004$ & $0.000\!\pm\!0.003$ \\
\hline
\end{tabular*}
\par\smallskip
\parbox{\linewidth}{\footnotesize
Regrets are normalized and averaged equally over nine decision thresholds.
For each regret component, $\Delta R=R(p_{\mathrm{emp}})-R(p^\star)$.
$\Delta\mathrm{MSE}=\mathrm{MSE}(\hat p,p_{\mathrm{emp}})-\mathrm{MSE}(\hat p,p^\star)$.
Entries show the estimate $\pm$ the larger distance to either endpoint of the
95\% stratified paired bootstrap interval, rounded outward.}
\end{table}

\begin{table}[t]
\centering
\caption{Changes in regret and stated-belief error under empirical frequency versus the true posterior ($d=5$).}
\label{tab:posterior-frequency-comparison-d5}
\scriptsize
\setlength{\tabcolsep}{1pt}
\begin{tabular*}{\linewidth}{@{\extracolsep{\fill}}lrrrr@{}}
\hline
Model / reasoning mode & $\Delta R_{\mathrm{total}}$ & $\Delta R_{\mathrm{belief}}$ & $\Delta R_{\mathrm{opt}}$ & $\Delta\mathrm{MSE}$ \\
\hline
Claude Haiku 4.5 / high & $-0.017\!\pm\!0.005$ & $-0.014\!\pm\!0.005$ & $-0.002\!\pm\!0.005$ & $-0.021\!\pm\!0.007$ \\
Claude Haiku 4.5 / none & $-0.005\!\pm\!0.007$ & $-0.009\!\pm\!0.008$ & $+0.004\!\pm\!0.007$ & $-0.017\!\pm\!0.010$ \\
Claude Sonnet 4.6 / high & $-0.019\!\pm\!0.004$ & $-0.012\!\pm\!0.005$ & $-0.007\!\pm\!0.004$ & $-0.014\!\pm\!0.005$ \\
Claude Sonnet 4.6 / none & $-0.012\!\pm\!0.005$ & $-0.007\!\pm\!0.005$ & $-0.005\!\pm\!0.004$ & $-0.006\!\pm\!0.005$ \\
Gemini 3.1 Flash Lite / high & $-0.017\!\pm\!0.004$ & $-0.019\!\pm\!0.005$ & $+0.002\!\pm\!0.004$ & $-0.028\!\pm\!0.007$ \\
Gemini 3.1 Flash Lite / minimal & $+0.009\!\pm\!0.010$ & $+0.015\!\pm\!0.007$ & $-0.007\!\pm\!0.007$ & $+0.017\!\pm\!0.007$ \\
Gemini 3.1 Pro / high & $-0.016\!\pm\!0.005$ & $-0.007\!\pm\!0.007$ & $-0.009\!\pm\!0.005$ & $-0.010\!\pm\!0.008$ \\
Gemini 3.1 Pro / low & $-0.017\!\pm\!0.005$ & $-0.011\!\pm\!0.005$ & $-0.006\!\pm\!0.004$ & $-0.014\!\pm\!0.006$ \\
Gemini 3.5 Flash / high & $-0.012\!\pm\!0.006$ & $-0.002\!\pm\!0.010$ & $-0.010\!\pm\!0.009$ & $-0.002\!\pm\!0.013$ \\
Gemini 3.5 Flash / minimal & $-0.013\!\pm\!0.005$ & $-0.011\!\pm\!0.004$ & $-0.002\!\pm\!0.004$ & $-0.013\!\pm\!0.004$ \\
Gemini 3 Flash / high & $-0.019\!\pm\!0.005$ & $-0.013\!\pm\!0.007$ & $-0.006\!\pm\!0.007$ & $-0.017\!\pm\!0.008$ \\
Gemini 3 Flash / minimal & $-0.010\!\pm\!0.006$ & $-0.012\!\pm\!0.006$ & $+0.002\!\pm\!0.007$ & $-0.018\!\pm\!0.007$ \\
GPT-5.4 Mini / medium & $-0.016\!\pm\!0.005$ & $-0.018\!\pm\!0.006$ & $+0.001\!\pm\!0.005$ & $-0.027\!\pm\!0.008$ \\
GPT-5.4 Mini / none & $+0.012\!\pm\!0.008$ & $+0.020\!\pm\!0.008$ & $-0.008\!\pm\!0.007$ & $+0.023\!\pm\!0.009$ \\
GPT-5.5 / medium & $-0.021\!\pm\!0.005$ & $-0.014\!\pm\!0.005$ & $-0.008\!\pm\!0.005$ & $-0.016\!\pm\!0.006$ \\
GPT-5.5 / none & $-0.010\!\pm\!0.005$ & $-0.009\!\pm\!0.005$ & $-0.002\!\pm\!0.006$ & $-0.012\!\pm\!0.005$ \\
Llama 4 Scout / thinking & $+0.016\!\pm\!0.009$ & $+0.007\!\pm\!0.009$ & $+0.009\!\pm\!0.011$ & $+0.005\!\pm\!0.012$ \\
Qwen 3.6 35B-A3B / none & $+0.020\!\pm\!0.010$ & $+0.023\!\pm\!0.010$ & $-0.002\!\pm\!0.009$ & $+0.027\!\pm\!0.013$ \\
\hline
\end{tabular*}
\par\smallskip
\parbox{\linewidth}{\footnotesize
Regrets are normalized and averaged equally over nine decision thresholds.
For each regret component, $\Delta R=R(p_{\mathrm{emp}})-R(p^\star)$.
$\Delta\mathrm{MSE}=\mathrm{MSE}(\hat p,p_{\mathrm{emp}})-\mathrm{MSE}(\hat p,p^\star)$.
Entries show the estimate $\pm$ the larger distance to either endpoint of the
95\% stratified paired bootstrap interval, rounded outward.}
\end{table}

\clearpage

\section{Generalization to unseen thresholds and decision structures}
\label{app:utility_generalization}
\begin{table}[ht]
  \centering

  \resizebox{\linewidth}{!}{%
    \begin{tabular}{lcccc}
    \hline
    \multicolumn{1}{c}{\textbf{Strategy}} & \multicolumn{1}{c}{\textbf{Format}} & \multicolumn{1}{c}{\textbf{Trained thresholds}} & \multicolumn{1}{c}{\textbf{Untrained thresholds}} & \multicolumn{1}{c}{\textbf{Gap (untrained $-$ trained)}} \\ \hline
    B     & independent & $-0.037^{*}\ [-0.053,\,-0.021]$ & $-0.033^{*}\ [-0.049,\,-0.018]$ & $0.004\ [-0.012,\,0.020]$ \\
    D     & independent & $0.001\ [-0.013,\,0.015]$ & $-0.010\ [-0.024,\,0.005]$ & $-0.010\ [-0.028,\,0.007]$ \\
    B/D   & independent & $-0.027^{*}\ [-0.046,\,-0.008]$ & $-0.025^{*}\ [-0.045,\,-0.005]$ & $0.002\ [-0.015,\,0.021]$ \\
    B+A   & independent & $-0.067^{*}\ [-0.083,\,-0.052]$ & $-0.070^{*}\ [-0.086,\,-0.054]$ & $-0.003\ [-0.018,\,0.013]$ \\
    B     & two-turn    & $-0.040^{*}\ [-0.051,\,-0.030]$ & $-0.048^{*}\ [-0.060,\,-0.037]$ & $-0.008^{*}\ [-0.012,\,-0.004]$ \\
    D     & two-turn    & $-0.006\ [-0.016,\,0.003]$ & $-0.008\ [-0.018,\,0.002]$ & $-0.002\ [-0.006,\,0.003]$ \\
    B/D   & two-turn    & $-0.034^{*}\ [-0.045,\,-0.023]$ & $-0.046^{*}\ [-0.057,\,-0.035]$ & $-0.012^{*}\ [-0.016,\,-0.007]$ \\
    B+A   & two-turn    & $-0.035^{*}\ [-0.047,\,-0.025]$ & $-0.047^{*}\ [-0.059,\,-0.036]$ & $-0.011^{*}\ [-0.015,\,-0.007]$ \\ \hline
    \end{tabular}%
  }
  \caption{Change in total regret relative to Base on the Synthetic task, evaluated separately at thresholds seen during training ($0.2,0.4,0.6,0.8$) and held-out thresholds ($0.1,0.3,0.5,0.7,0.9$). The gap is the held-out value minus the trained-threshold value. Entries are means over 200 cases with 95\% paired bootstrap intervals; $^{*}$ marks intervals excluding zero.}
  \label{tab:r_syn_utility_generalization}
\end{table}

\begin{table}[ht]
\centering
\small
\setlength{\tabcolsep}{5pt}
\begin{tabular}{lllccc}
\toprule
\textbf{Model} & \textbf{Strategy} & \textbf{Format} & $\Delta R_{\mathrm{total}}$ & $\Delta R_{\mathrm{belief}}$ & $\Delta R_{\mathrm{opt}}$ \\
\midrule
Qwen3-8B & B & independent & $-0.036 \pm 0.014$ & $-0.026 \pm 0.006$ & $-0.010 \pm 0.015$ \\
 & B & two-turn & $-0.052 \pm 0.013$ & $-0.026 \pm 0.006$ & $-0.026 \pm 0.013$ \\
 & D & independent & $-0.123 \pm 0.019$ & $-0.013 \pm 0.004$ & $-0.110 \pm 0.019$ \\
 & D & two-turn & $-0.053 \pm 0.013$ & $-0.013 \pm 0.004$ & $-0.040 \pm 0.012$ \\
 & B/D & independent & $-0.120 \pm 0.019$ & $-0.025 \pm 0.006$ & $-0.095 \pm 0.019$ \\
 & B/D & two-turn & $-0.064 \pm 0.015$ & $-0.025 \pm 0.006$ & $-0.039 \pm 0.013$ \\
 & B+A & independent & $-0.039 \pm 0.013$ & $-0.027 \pm 0.005$ & $-0.012 \pm 0.014$ \\
 & B+A & two-turn & $-0.060 \pm 0.014$ & $-0.027 \pm 0.005$ & $-0.033 \pm 0.014$ \\
\midrule
Llama-3.1-8B & B & independent & $-0.006 \pm 0.011$ & $-0.016 \pm 0.013$ & $0.010 \pm 0.017$ \\
 & B & two-turn & $-0.009 \pm 0.010$ & $-0.016 \pm 0.013$ & $0.007 \pm 0.014$ \\
 & D & independent & $-0.072 \pm 0.016$ & $-0.022 \pm 0.014$ & $-0.050 \pm 0.023$ \\
 & D & two-turn & $-0.020 \pm 0.011$ & $-0.022 \pm 0.014$ & $0.001 \pm 0.016$ \\
 & B/D & independent & $-0.085 \pm 0.016$ & $-0.022 \pm 0.014$ & $-0.064 \pm 0.022$ \\
 & B/D & two-turn & $-0.023 \pm 0.010$ & $-0.022 \pm 0.014$ & $-0.001 \pm 0.016$ \\
 & B+A & independent & $0.003 \pm 0.010$ & $-0.016 \pm 0.014$ & $0.019 \pm 0.016$ \\
 & B+A & two-turn & $-0.013 \pm 0.010$ & $-0.016 \pm 0.014$ & $0.003 \pm 0.015$ \\
\bottomrule
\end{tabular}
\caption{Generalization to three-action clinical decision problems with a defer option. We report mean changes relative to the corresponding base model with 95\% bootstrap CI.}
\label{tab:three-action-generalization}
\end{table}

\clearpage
\section{Replicating RL interventions on Llama3.1-8B}
\label{app:replication}

\begin{figure}[htb]
    \centering
    \includegraphics[width=\textwidth]{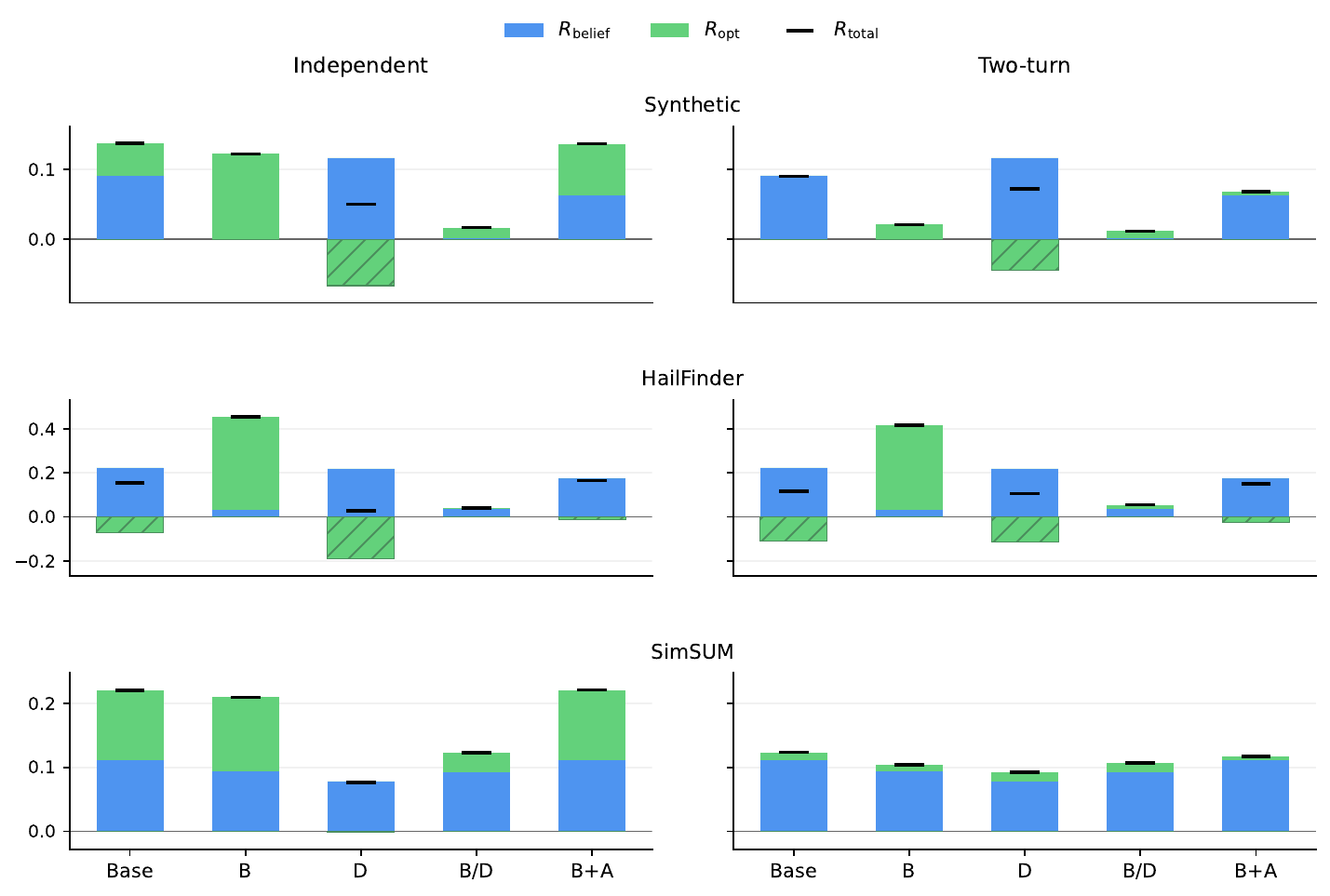}
    \caption{Aggregated regret decomposition of the base model (Llama3.1-8B) and after the four interventions. Rows show synthetic, weather and clinical tasks; and columns show independent and two-turn evaluation.}
    \label{fig:result_abs_llama}
\end{figure}

Figure~\ref{fig:result_abs_llama} show the aggregated performance of the base model (Llama3.1-8B) and after each of the four interventions. 

\begin{figure}[htb]
    \centering
    \includegraphics[width=\textwidth]{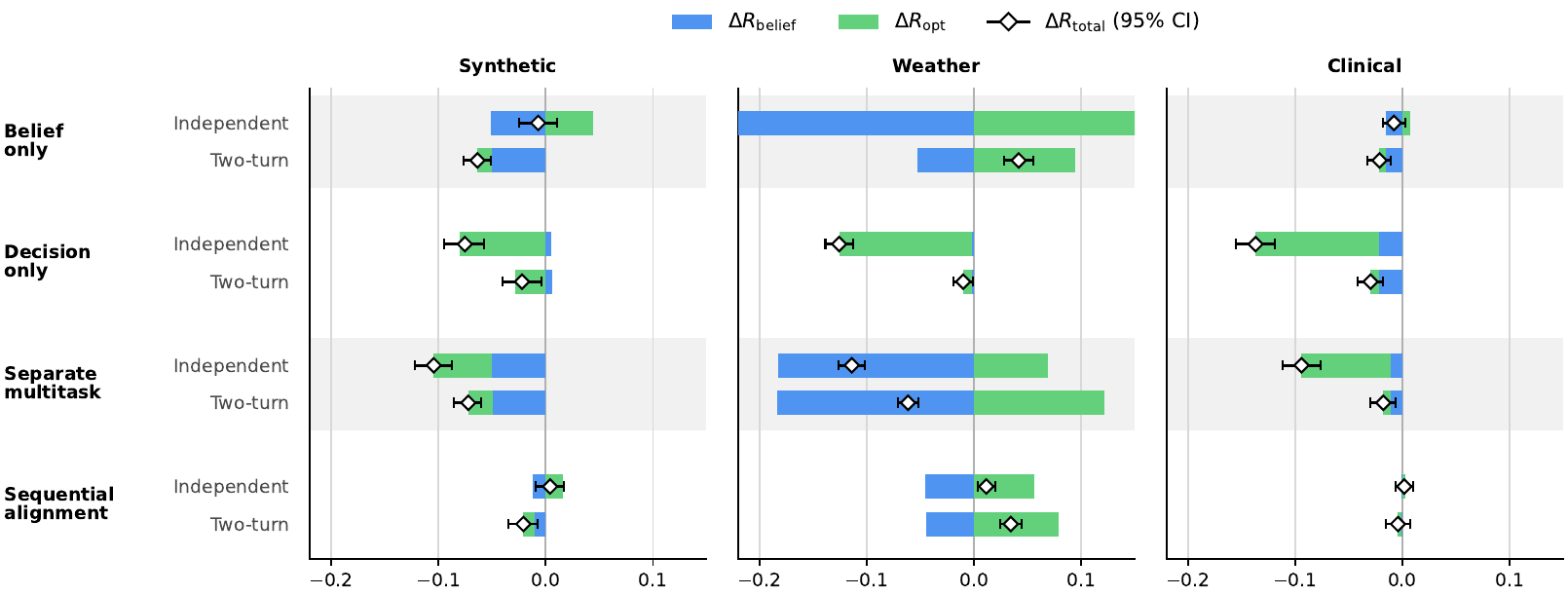}
    \caption{Averaged differences from the base model (Llama3.1-8B) in normalized total regret, belief loss, and optimization residual. Rows show the four interventions (belief-only, decision-only, separate multitask, and sequential alignment training); and subrows distinguish between the independent and two-turn evaluation formats. Values below zero indicate a reduction (improvement) in the corresponding term relative to the base model.}
    \label{fig:result_diff_llama}
\end{figure}

Figure~\ref{fig:result_diff_llama} show the effects of each intervention on the regret decomposition. We provide numerical results with 95\% bootstrap confidence intervals in the following Table~\ref{tab:llama-effects-synthetic}, ~\ref{tab:llama-effects-weather}, ~\ref{tab:llama-effects-clinical}.

% Synthetic
\begin{table}[ht]
\centering
\small
    \begin{tabular}{llccccc}
    \toprule
    \textbf{Strategy} & \textbf{Format} & $\Delta R_{\mathrm{total}}$ & $\Delta R_{\mathrm{belief}}$ & $\Delta R_{\mathrm{opt}}$ & $\Delta \mathrm{Belief\_MSE}$ & $\Delta \mathrm{Agreement}$ \\
    \midrule
    B & independent & $-0.007 \pm 0.018$ & $-0.051 \pm 0.011$ & $0.044 \pm 0.021$ & $-0.061 \pm 0.012$ & $-0.002 \pm 0.020$ \\
    B & two-turn & $-0.064 \pm 0.013$ & $-0.050 \pm 0.010$ & $-0.014 \pm 0.011$ & $-0.061 \pm 0.012$ & $-0.040 \pm 0.014$ \\
    D & independent & $-0.075 \pm 0.018$ & $0.005 \pm 0.016$ & $-0.080 \pm 0.026$ & $0.009 \pm 0.020$ & $-0.060 \pm 0.019$ \\
    D & two-turn & $-0.022 \pm 0.018$ & $0.006 \pm 0.016$ & $-0.028 \pm 0.013$ & $0.009 \pm 0.020$ & $-0.045 \pm 0.018$ \\
    B/D & independent & $-0.104 \pm 0.017$ & $-0.050 \pm 0.011$ & $-0.054 \pm 0.019$ & $-0.060 \pm 0.012$ & $-0.103 \pm 0.015$ \\
    B/D & two-turn & $-0.072 \pm 0.013$ & $-0.049 \pm 0.010$ & $-0.023 \pm 0.011$ & $-0.060 \pm 0.012$ & $-0.051 \pm 0.014$ \\
    B+A & independent & $0.004 \pm 0.013$ & $-0.012 \pm 0.011$ & $0.016 \pm 0.017$ & $-0.013 \pm 0.014$ & $0.004 \pm 0.016$ \\
    B+A & two-turn & $-0.021 \pm 0.014$ & $-0.010 \pm 0.011$ & $-0.010 \pm 0.012$ & $-0.013 \pm 0.014$ & $-0.021 \pm 0.016$ \\
    \bottomrule
    \end{tabular}
\caption{Synthetic. Changes from the Llama-3.1-8B baseline. Mean difference with 95\% confidence interval (n = 200 cases). Negative values indicate lower regret than Base.}
\label{tab:llama-effects-synthetic}
\end{table}

% Weather
\begin{table}[ht]
\centering
\small
    \begin{tabular}{llccccc}
    \toprule
    \textbf{Strategy} & \textbf{Format} & $\Delta R_{\mathrm{total}}$ & $\Delta R_{\mathrm{belief}}$ & $\Delta R_{\mathrm{opt}}$ & $\Delta \mathrm{Belief\_MSE}$ & $\Delta \mathrm{Agreement}$ \\
    \midrule
    B & independent & $0.228 \pm 0.051$ & $-0.466 \pm 0.051$ & $0.694 \pm 0.081$ & $-0.305 \pm 0.033$ & $0.386 \pm 0.043$ \\
    B & two-turn & $0.042 \pm 0.014$ & $-0.053 \pm 0.014$ & $0.095 \pm 0.024$ & $-0.226 \pm 0.019$ & $-0.087 \pm 0.022$ \\
    D & independent & $-0.126 \pm 0.013$ & $-0.002 \pm 0.005$ & $-0.124 \pm 0.014$ & $-0.003 \pm 0.006$ & $0.102 \pm 0.017$ \\
    D & two-turn & $-0.010 \pm 0.009$ & $-0.002 \pm 0.005$ & $-0.008 \pm 0.009$ & $-0.003 \pm 0.006$ & $-0.004 \pm 0.010$ \\
    B/D & independent & $-0.114 \pm 0.012$ & $-0.183 \pm 0.012$ & $0.069 \pm 0.012$ & $-0.218 \pm 0.014$ & $-0.159 \pm 0.012$ \\
    B/D & two-turn & $-0.062 \pm 0.010$ & $-0.183 \pm 0.012$ & $0.122 \pm 0.010$ & $-0.218 \pm 0.014$ & $-0.076 \pm 0.010$ \\
    B+A & independent & $0.011 \pm 0.008$ & $-0.045 \pm 0.006$ & $0.057 \pm 0.010$ & $-0.060 \pm 0.007$ & $-0.055 \pm 0.010$ \\
    B+A & two-turn & $0.034 \pm 0.010$ & $-0.045 \pm 0.006$ & $0.079 \pm 0.010$ & $-0.060 \pm 0.007$ & $-0.099 \pm 0.010$ \\
    \bottomrule
    \end{tabular}
\caption{Weather. Changes from the Llama-3.1-8B baseline. Mean difference with 95\% confidence interval (n = 500 cases). Negative values indicate lower regret than Base.}
\label{tab:llama-effects-weather}
\end{table}

% Clinical
\begin{table}[ht]
\centering
\small
    \begin{tabular}{llccccc}
    \toprule
    \textbf{Strategy} & \textbf{Format} & $\Delta R_{\mathrm{total}}$ & $\Delta R_{\mathrm{belief}}$ & $\Delta R_{\mathrm{opt}}$ & $\Delta \mathrm{Belief\_MSE}$ & $\Delta \mathrm{Agreement}$ \\
    \midrule
    B & independent & $-0.008 \pm 0.011$ & $-0.015 \pm 0.010$ & $0.007 \pm 0.014$ & $-0.019 \pm 0.012$ & $-0.010 \pm 0.012$ \\
    B & two-turn & $-0.021 \pm 0.011$ & $-0.015 \pm 0.010$ & $-0.006 \pm 0.008$ & $-0.019 \pm 0.012$ & $-0.001 \pm 0.006$ \\
    D & independent & $-0.137 \pm 0.018$ & $-0.022 \pm 0.011$ & $-0.115 \pm 0.020$ & $-0.028 \pm 0.012$ & $-0.065 \pm 0.015$ \\
    D & two-turn & $-0.030 \pm 0.012$ & $-0.022 \pm 0.011$ & $-0.008 \pm 0.009$ & $-0.028 \pm 0.013$ & $-0.011 \pm 0.006$ \\
    B/D & independent & $-0.094 \pm 0.018$ & $-0.010 \pm 0.010$ & $-0.084 \pm 0.020$ & $-0.016 \pm 0.012$ & $-0.043 \pm 0.014$ \\
    B/D & two-turn & $-0.018 \pm 0.012$ & $-0.011 \pm 0.010$ & $-0.007 \pm 0.009$ & $-0.016 \pm 0.012$ & $-0.007 \pm 0.006$ \\
    B+A & independent & $0.002 \pm 0.008$ & $-0.001 \pm 0.010$ & $0.003 \pm 0.014$ & $-0.002 \pm 0.012$ & $0.000 \pm 0.011$ \\
    B+A & two-turn & $-0.004 \pm 0.011$ & $-0.001 \pm 0.011$ & $-0.003 \pm 0.009$ & $-0.002 \pm 0.012$ & $-0.005 \pm 0.006$ \\
    \bottomrule
    \end{tabular}
\caption{Clinical. Changes from the Llama-3.1-8B baseline. Mean difference with 95\% confidence interval (n = 500 cases). Negative values indicate lower regret than Base.}
\label{tab:llama-effects-clinical}
\end{table}

\section{Genuine optimization errors vs. belief-action misalignment}
\label{app:distinguish_R_opt}

To answer whether measured optimization residuals mainly reflect genuine optimization errors or belief-action misalignment, we add a supplementary experiment to evaluate the decision consequences when a model is directly provided with the beliefs, and must only choose the best action. We compare three conditions: 1) independent evaluation, which elicits beliefs and actions in separate sessions; 2) two-turn evaluation, which gives the model access to evidence and its reported beliefs; and 3) belief-only evaluation, which directly supplies the reported beliefs to the model and asks for actions without the original evidence. We evaluate mean optimization residuals and agreement between belief-implied and reported actions. We test GPT-5.5 in both thinking and non-thinking modes and provide the results in the following table.

We find that with medium thinking effort, the model reliably chooses belief-implied optimal actions when probabilities are explicitly available. This suggests that in this setting, optimization residuals in independent evaluation may mainly reflect belief-action misalignment differences between the beliefs used to choose actions and those reported, rather than a general inability to optimize under supplied probabilities. On the other hand, we find that without thinking, belief–action agreement remains between 87.0\% and 89.9\% across the three conditions. This may reflect genuine difficulty optimizing the given utility functions, although we cannot rule out the possibility that the model does not fully adopt the supplied beliefs. Table~\ref{tab:gpt55-belief-action} summarizes the results.

\begin{table}[t]
\centering
\small
\setlength{\tabcolsep}{5pt}
\begin{tabular}{llcc}
\toprule
\textbf{Thinking mode} & \textbf{Evaluation format} &
\textbf{Mean} $\mathrm{R_opt}$ &
\shortstack{\textbf{Belief--action}\\\textbf{agreement}} \\
\midrule
None & Independent & 0.018 & 87.0\% \\
None & Two-turn & 0.023 & 89.9\% \\
None & Belief-only & 0.022 & 87.7\% \\
\midrule
Medium & Independent & 0.0043 & 98.4\% \\
Medium & Two-turn & 0.0003 & 99.8\% \\
Medium & Belief-only & 0.0003 & 100\% \\
\bottomrule
\end{tabular}
\caption{Mean optimization residual and belief--action agreement for GPT-5.5 across thinking modes and evaluation formats.}
\label{tab:gpt55-belief-action}
\end{table}

\end{document}